\documentclass{article}

\PassOptionsToPackage{numbers, compress}{natbib}
\usepackage[preprint]{neurips_2025}

\usepackage[utf8]{inputenc} % allow utf-8 input
\usepackage[T1]{fontenc}    % use 8-bit T1 fonts
\usepackage{xcolor}         % colors
\definecolor{mycitecolor}{RGB}{0,0,127} % 深红色
\usepackage[colorlinks=true, linkcolor=mycitecolor, urlcolor=mycitecolor, citecolor=mycitecolor]{hyperref}       % hyperlinks
\usepackage{url}            % simple URL typesetting
\usepackage{booktabs}       % professional-quality tables
\usepackage{amsfonts}       % blackboard math symbols
\usepackage{nicefrac}       % compact symbols for 1/2, etc.
\usepackage{microtype}      % microtypography

\usepackage{float}
\usepackage{pifont}
\usepackage{wrapfig}
\usepackage{amsmath,amssymb,amsfonts,bm}
\usepackage{algorithmic}
\usepackage{graphicx}
\usepackage{textcomp}
\usepackage{siunitx}
\usepackage{soul}
\usepackage{xspace}
\usepackage{booktabs}
\usepackage{subfig}
\usepackage{multirow}
\usepackage{caption}
\usepackage{balance}
\usepackage[inline]{enumitem}
\usepackage[export]{adjustbox}
\usepackage[ruled,vlined]{algorithm2e}
\usepackage{marvosym} %符号的由来

\newcommand{\etc}{\emph{etc.}\xspace}

\newcommand{\figref}[1]{\hyperref[#1]{Figure~\ref*{#1}}}
\newcommand{\tabref}[1]{\hyperref[#1]{Table~\ref*{#1}}}
\newcommand{\secref}[1]{\hyperref[#1]{\S~\ref*{#1}}}
\newcommand{\appendixref}[1]{\hyperref[#1]{Appendix~\ref*{#1}}}
\newcommand{\equref}[1]{\hyperref[#1]{Equation~\ref*{#1}}}

\newcommand{\methodname}{{SpecOffload}\xspace}
\newcommand{\squishlist}{
\begin{list}{$\bullet$}{
  \setlength{\itemsep}{0pt}
  \setlength{\parsep}{3pt}
  \setlength{\topsep}{3pt}
  \setlength{\partopsep}{0pt}
  \setlength{\leftmargin}{3.5mm}
  \setlength{\labelwidth}{1em}
  \setlength{\labelsep}{0.5em}}}
\newcommand{\squishend}{\end{list}}

\title{SpecOffload: Unlocking Latent GPU Capacity for  LLM Inference on Resource-Constrained Devices }

\author{%
  Xiangwen Zhuge$^{1}$, 
  Shen Xu$^{1}$, 
  Zeyu Wang$^{1}$, 
  Fan  Dang$^{2}$, 
  Xuan Ding$^{1}$, 
  Dangyang Li$^{1}$, \\
  \textbf{Yahui Han}$^{3}$, 
  \textbf{Tianxiang Hao}$^{1}$, 
  \textbf{Zheng Yang}$^{1}$ \\
  \\
  $^1$Tsinghua University \\
  $^2$Beijing Jiaotong University \\
  $^3$Beijing University of Posts and Telecommunications
}

\begin{document}

\maketitle

\begin{abstract}
%   \TODO{[GPT write]}  The evolution of Large Language Models (LLMs) has expanded their applications to a broader range of scenarios, such as throughput-oriented data generation tasks, while the expansion of model parameters has imposed higher demands on GPU resources. To enable LLM deployment on platforms with limited GPU resources, this paper presents \methodname, an Offloading framework for high-throughput LLM inference. Our analysis reveals two major deficiencies in SOTA offloading methods: underutilization of GPU cores and marginal utility of GPU memory (VRAM). To address these issues, we delicately embedded SD into Offloading at virtually zero cost, utilizing idle GPU computational resources for draft model operations and reallocating inefficiently used VRAM for storing draft model parameters. Our system achieves efficient integration of heterogeneous models and computational resources through Adaptive Tensor Placement, ParaSpec Planner, and Interleaved Batch pipeline. Experimental results demonstrate that our approach achieves throughput improvements of 2.51× over best SOTA system, while increasing GPU core utilization by 4.49 times. Our code is available at \href{www.baidu.com}{www.baidu.com}.

Efficient LLM inference on resource-constrained devices presents significant challenges in compute and memory utilization. Due to limited GPU memory, existing systems offload model weights to CPU memory, incurring substantial I/O overhead between the CPU and GPU. This leads to two major inefficiencies: (1) GPU cores are underutilized, often remaining idle while waiting for data to be loaded; and (2) GPU memory has a low impact on performance, as reducing its capacity has minimal effect on overall throughput.
In this paper, we propose \methodname, a high-throughput inference engine that embeds speculative decoding into offloading. Our key idea is to unlock latent GPU resources for storing and executing a draft model used for speculative decoding, thus accelerating inference at near-zero additional cost. To support this, we carefully orchestrate the interleaved execution of target and draft models in speculative decoding within the offloading pipeline, and propose a planner to manage tensor placement and select optimal parameters. Compared with the best baseline, \methodname improves GPU core utilization by 4.49× and boosts inference throughput by 2.54×. Our code is available at \href{https://github.com/MobiSense/SpecOffload-public}{https://github.com/MobiSense/SpecOffload-public}.
\end{abstract}

% This leads to two major inefficiencies。这里或之前缺了关键的“IO”与GPU算力不平衡这一本质问题，对应你incurring substantial data transfer overhead between the CPU and GPU这一句，这里建议体现一下IO这个词就ok了。

% (1) underutilization of GPU cores, which often remain idle while waiting for data to be loaded; and (2) marginal utility of GPU memory, as its limited capacity means further reductions have little effect on throughput.这两句主语、语法啥的没问题吧。which是指的GPU我懂，但整体读起来有点怪。

%  marginal utility of GPU memory, as its limited capacity means further reductions have little effect on throughput. 这句我觉得reviewer看不懂。这里直接说现象吧，增加或减少用于暂存（加载）LLM的显存对吞吐量影响十分有限。

% Building on these observations，这个衔接0分。不如，In this paper，xxx。你要说Observation就要说思路，而不是提出xxx。例如，我观察到GPU空着，那我就要想办法填满它，进而退出sysname。

% a high-throughput inference “engine”？or framework？
% that embeds speculative decoding mechanism into the offloading paradigm.

\section{Introduction}
As Large Language Models (LLMs) evolve, their real-world use extends far beyond chatbots to diverse applications including synthetic data generation \cite{datageneration}, form processing \cite{formprocessing}, and data wrangling \cite{datawrangling}. These tasks are characterised by LLMs conducting offline inference in batches over a large number of tokens. For instance, corporations need to process all archives of financial documentation, whilst individuals want to construct knowledge repositories from accumulated materials. In such workloads, higher inference throughput (the number of tokens generated divided by total generation time, token/s) translates into lower total completion time, hence it is the key metric.

Privacy and cost concerns drive these tasks toward LLM deployment on edge servers or PCs, where GPU memory is a major constraint. For instance, conducting inference exclusively using GPU memory for Mixtral 8x22B (282GB) \cite{Mixtral8x22b} requires at least four NVIDIA H100 (80GB) GPUs. Offloading is one of the mainstream solutions to memory-constrained inference, transferring most model parameters to more economical, capacious CPU memory and reloading them to GPU memory only when computation demands. There are also methods to overcome the memory bottleneck by compressing the model and KV cache, such as quantization, pruning, sparsification \cite{quantization, pruning, sparsification}, 
\etc, which are orthogonal and can be applied on top of offloading. Our focus is on designing efficient offloading strategies for high-throughput inference on resource-constrained devices.
% for single GPU? 不需要cpu了？
% 别人也是这么写的.jpg

However, the existing offloading work does not utilize GPU resources well. During offloading, generating each token requires reloading most model parameters from CPU memory to GPU memory for execution \cite{deepspeed, fastinference_offloading, Klotski}. Yet I/O speeds substantially lag behind GPU computational capabilities. For instance, under typical NVIDIA RTX 4090, PCIe 4.0×16 conditions, loading a single FFN layer of the Mixtral 8×22B decoder from CPU to GPU consumes 240ms, while the actual computation on GPU requires merely 0.1ms. Consequently, total inference time is mainly determined by parameter loading time, leaving GPU resources severely underutilized.

\begin{figure}[tbp]
    \centering
    \begin{minipage}[b]{0.48\textwidth}
        \centering
        \includegraphics[width=\linewidth]{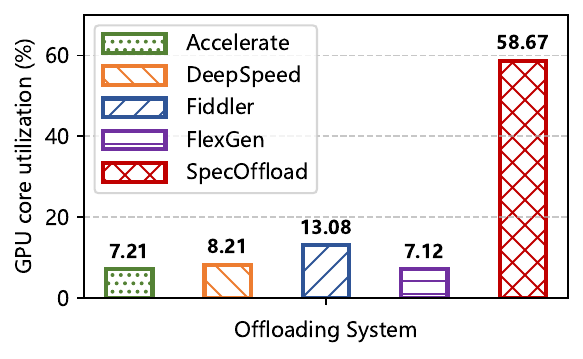}
        \caption{GPU core utilization of SOTA methods during decoding phase. Settings: Mixtral 8x7B, Env \#1, SummEval dataset, details in  \secref{sec:expsetup}.}
        \label{fig:motivation1}
    \end{minipage}
    \hfill
    \begin{minipage}[b]{0.48\textwidth}
        \centering
        \includegraphics[width=\linewidth]{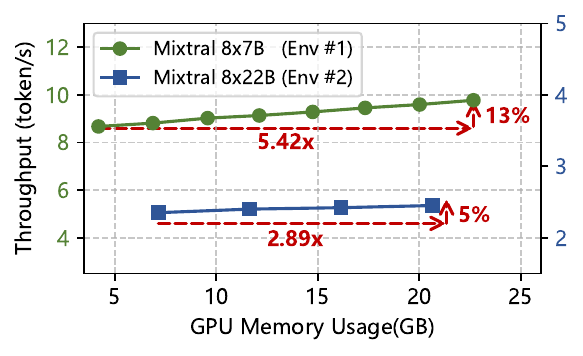}
        \caption{Impact of GPU memory on throughput during decoding phase. Settings: SummEval dataset, details in \secref{sec:expsetup}.}
        \label{fig:motivation2}
    \end{minipage}
    \vspace{-1.5em}
\end{figure}

To highlight the inefficiencies of existing approaches, we perform a detailed analysis of GPU core and memory utilization. We find that:

\begin{itemize}[leftmargin=2em, topsep=-2pt, parsep=0pt]
    \item \textbf{Underutilization of GPU cores}. As shown in \figref{fig:motivation1}, during the decoding phase, the average GPU core utilization of existing methods is only 13\% at most.
    % \TODO{ (Experimental settings are detailed in \secref{sec:expsetup}.)}
    This inefficiency stems from GPU cores frequently remaining long time idle while awaiting parameter loading. To alleviate this issue, existing methods increase the batch size to amortize I/O overhead by loading model parameters once and reusing them across the enlarged batch, thereby improving throughput \cite{Flexgen,Klotski}. However, due to the limitation of GPU memory capacity or CPU computational capabilities, the scalability of the batch size remains inherently limited. The maximum batch size achieved by the SOTA scheme in \figref{fig:motivation1} is only 64, insufficient to bridge the huge gap between I/O and GPU latency (even with this batch size, the gap remains over 10×).

    \item \textbf{Marginal utility of GPU memory}. When models far exceed GPU memory capacity, throughput remains almost unchanged even when memory usage is greatly reduced during decoding phase. We conducted experiments with FlexGen \cite{Flexgen}, the most effective solution among SOTA approaches. FlexGen offloads attention computations to CPU while computing FFN layers on GPU during the decoding phase.
    % \TODO{Experimental settings are detailed in \secref{sec:expsetup}.}
    As illustrated in \figref{fig:motivation2}, reducing memory usage by over 5.42× for Mixtral 8×7B models led to only a 13\% drop in throughput; similarly, a 2.89× reduction for Mixtral 8×22B models caused just a 5\% decline. This phenomenon arises because the total volume of data to be loaded remains nearly constant, as the model size vastly exceeds the available GPU memory, leaving little room for variation even when accounting for partical parameters that can reside permanently in memory. For instance, generating a single token necessitates loading all (56) FFN layers for the leftmost blue point in \figref{fig:motivation2}, compared to 53 layers for the rightmost blue point. Consequently, memory reduction precipitates only negligible changes in total inference time and throughput.
\end{itemize}

To harness GPU compute and memory resources more efficiently, we design \methodname, a novel offloading framework that unlocks latent GPU capacity by leveraging speculative decoding (SD).
SD is a technique that accelerates generation by employing an auxiliary lightweight draft model to produce multiple candidate tokens, which are subsequently verified in parallel by the target model, enabling the generation of multiple tokens per forward pass \cite{blockwise, FastInference, ACloseLook}.

\methodname embeds SD into the offloading workflow with nearly zero overhead. The key idea lies in the following two aspects:

\begin{itemize}[leftmargin=2em, topsep=-2pt, parsep=0pt]
    \item Computing draft model during GPU core idleness: SD requires the draft model to generate multiple candidate tokens in advance. Given the substantial idle periods prevalent in existing frameworks, these intervals can be utilized for completing the draft model's computational tasks.
    \item Storing draft model uses "low-yield" GPU memory: SD requires loading a draft model into memory for draft generation. We can repurpose "low-yield" memory allocations from existing frameworks to store draft model parameters and its caches instead. For instance, as shown in \figref{fig:motivation2}, extracting 17GB of "low-yield" memory allocation provides sufficient capacity for a draft model such as Mistral 7B \cite{mistral7b} to operate normally within the GPU at a small batch.
\end{itemize}

% We have reorganized GPU computational and storage resources, enabling the process of SD to be delicately embedded into Offloading at virtually zero cost. In contrast to existing paradigms, where GPUs are dedicated solely to the computation and storage of the main model, our approach allocates both the computation and storage responsibilities of the draft model to the GPU.
To support this, \methodname designs a comprehensive framework to better utilize both the computational and memory resources of the GPU. \methodname determines tensor distribution between GPU and CPU memory through offline Adaptive Tensor Placement (\secref{sec:placement}), dynamically schedules computational tasks via the online ParaSpec Planner (\secref{sec:paraspec}), and implements parallel pipelined execution of I/O and computation using the Interleaved Batch Pipeline (\secref{sec:pipeline}).

Our contributions are as follows:

\begin{itemize}[leftmargin=2em, topsep=-2pt, parsep=0pt]
    \item We conduct a quantitative analysis of GPU resource utilization in representative scenarios and identify key limitations in SOTA frameworks—underutilization of GPU cores and marginal utility of GPU memory, thus reveal a novel perspective for enhancing offloading performance.
    
    \item  By designing a sub-layer model decomposition and fine-grained scheduling of  compute and memory resources, we delicately embedded SD into offloading with virtually zero overhead, thereby increasing GPU core utilization by 4.49 times. % where enhancing the cost-effectiveness of VRAM usage.
    % Through sub-layer level model decomposition and  pipeline design integrating heterogeneous models and computational resources,
    
    \item To evaluate \methodname, we benchmark it against HuggingFace Accelerate \cite{accelerate}, DeepSpeed-FastGen \cite{deepspeed}, FlexGen \cite{Flexgen}, and Fiddler \cite{fiddler}. Results demonstrate that our system achieves throughput improvements of 4.69×, 4.71×, 2.54×, and 4.04× over these baselines, respectively.
\end{itemize}

\section{Background and Related Work}

\subsection{Generative LLM Inference}
Large language models consist of stacked Transformer layers %, with weights primarily concentrated in the query/key/value projection matrices of the Attention and the linear transformation matrices of the feed-forward networks (FFN) 
\cite{attentionisallyouneed}. During inference, tokens are generated in an autoregressive paradigm: the prefill phase processes the complete input sequence to construct the KV cache, whereas in the decoding phase, subsequent token is produced based on the previously generated tokens and the cached KV states \cite{pageattention, pope_efficiently_2023, GPT3}.

\subsection{Speculative Decoding}
Speculative decoding (SD) is a method for accelerating LLM inference. It adheres to a “Draft-then-Verify” framework: at each decoding step, a lightweight draft model initially proposes multiple candidate tokens (\textit{e.g.}, $(\hat{w}_1, \hat{w}_2, \hat{w}_3, \hat{w}_4$)), which are collectively verified by a larger target model in a single forward pass. Only the valid subset $({w}_1, {w}_2)$ is accepted, after which the target model resumes decoding by independently generating the subsequent token $w_3$ \cite{blockwise, FastInference, ACloseLook}. This approach enables the target model to generate multiple tokens per inference step. To further enhance the efficiency of 
SD, prior research has predominantly explored two avenues: the design of more effective draft models \cite{distillspec, draft-verify, onlineSD} and draft structures \cite{medusa, specinfer, specexec}. 
% Speculative decoding introduces additional computational and memory overhead due to the inclusion of a draft model, so it is typically employed in scenarios with abundant GPU resources. While, in this work, we observe that GPU resources remain underutilized in offloading settings. To address this, we propose a fine-grained model partitioning strategy coupled with a heterogeneous pipelined execution framework, enabling the delicate integration of speculative decoding into the offloading process, which yields impressive performance gains.
%While SD increases GPU overhead and appears unsuitable for offloading because of using draft model, our fine-grained partitioning of model and heterogeneous pipeline enable near-zero-cost integration of the draft model, significantly boosting throughput.
However, traditional single-batch SD is not well-suited for integration with offloading, as the computations of the draft and target models must be executed sequentially. As a result, the GPU resources remain underutilized during the target model’s verification phase. By introducing a dual-batch rotation strategy, \methodname enables the verifying and drafting  to run concurrently, allowing SD to be seamlessly embedded into the offloading pipeline, while better utilizing GPU resources.

\subsection{Offloading in LLM Inference}
Offloading is one of the predominant solutions for enabling LLM inference under GPU memory constraints. It entails relocating certain model parameters  from the expensive, limited GPU memory to the more cost-effective and abundant CPU memory \cite{deepspeed,fiddler,powerinfer,moe-infinity,fastdecode,fastinference_offloading}. I/O constraints create the primary bottleneck in offloading, as data transfer latency between CPU memory and GPU memory far exceeds the GPU's computation time. Consequently, GPU resources remain underutilized.

In throughput-oriented scenarios, existing methods typically increase the batch size to amortize I/O costs. A prevalent strategy involves altering the model’s execution pattern from row-wise to column-wise, allowing each layer’s parameters to be loaded once and reused across multiple batches, thereby reducing the per-layer I/O burden \cite{Flexgen,Klotski}. However, this approach is constrained by the available GPU memory and the overhead 
I/O of KV cache. More recent studies demonstrate that offloading attention computation to the CPU can eliminate KV cache I/O in decoding phase \cite{moe-lightning,neo,hetegen,PACT'24}, but in doing so, CPU compute limitations cap batch scalability. Thus, GPU resources remain significantly underutilized. 
%However, by introducing SD, our approach effectively harnesses idle GPU computation and storage resources, thereby enhancing throughput.
In this work, by interleaving the draft model’s workload into the GPU’s idle periods between the target model’s layer-wise computations, we improve GPU core utilization by 4.49×.

\section{System Overview}

\begin{wrapfigure}{r}{0.48\textwidth}  % r 表示靠右，0.5\textwidth 表示宽度
    \centering
\includegraphics[width=\linewidth]{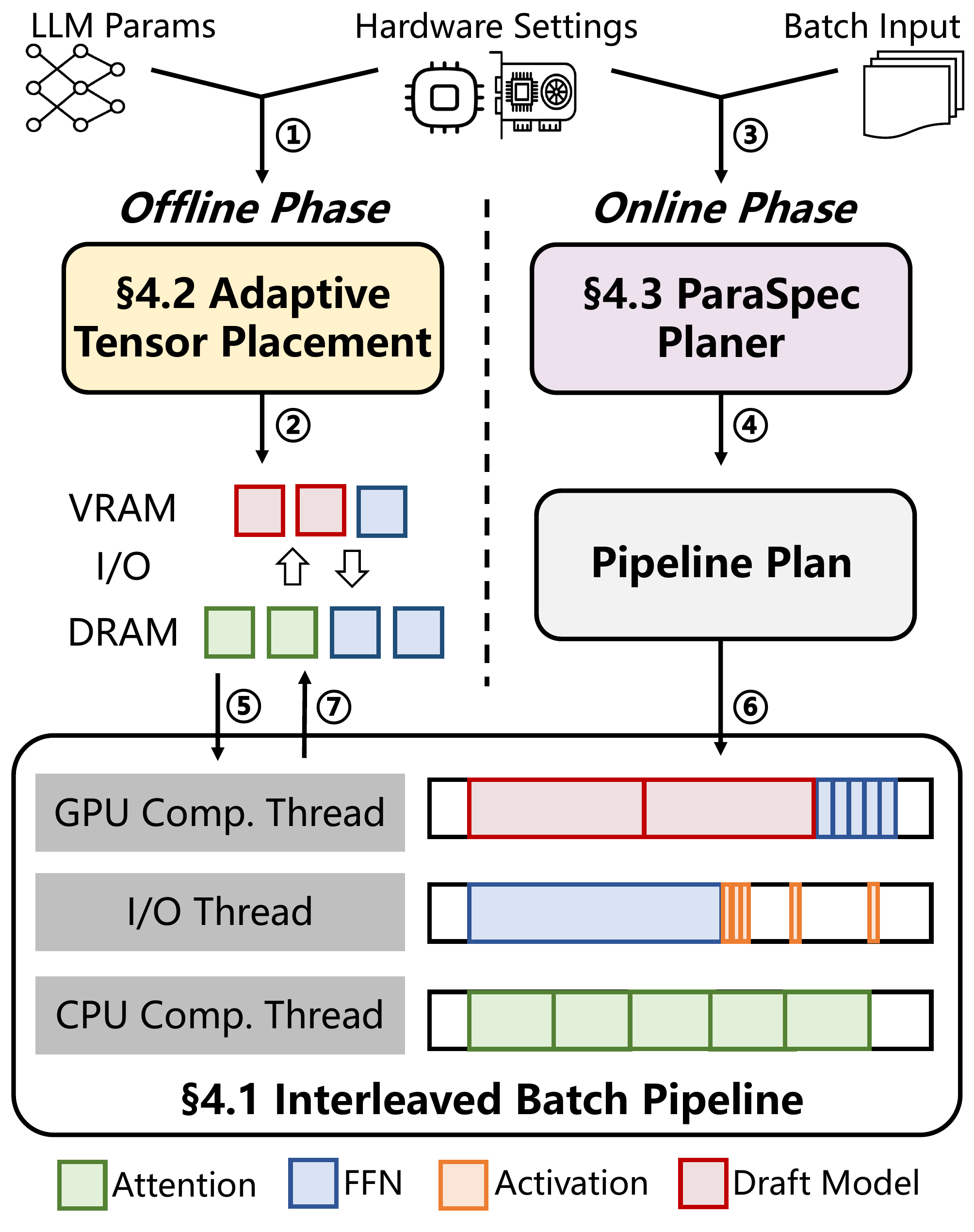}
    \caption{System overview of SpecOffload.}
    \label{fig:system-overview}
\end{wrapfigure}
\vspace{-10pt} % 减小 wrapfigure 之后的空白

% In this work, we propose \methodname, a high-throughput inference engine that embeds speculative decoding into offloading at near-zero additional cost. An architectural overview of \methodname is illustrated in \figref{fig:system-overview}.

% 能不能来点实在的，感觉这段话看了三遍了。就不能说说sysname有几个阶段，整体说一下，后面你突然说offline和online有点突兀

In this work, we propose \methodname. As shown in \figref{fig:system-overview}, it employs a two-phase architecture where offline  tensor placement and online scheduling collectively determine a unified pipeline across GPU and CPU.

During the offline phase, the target and draft models are deployed across a heterogeneous memory hierarchy. \methodname automatically evaluates the hardware ecosystem, measuring CPU and GPU memory capacities, computational performance of CPU and GPU cores, and the bandwidth of data transfer channels between them. These hardware specifications, along with the configurations of models (\ding{192}), are input to the Adaptive Tensor Placement (\secref{sec:placement}) to determine the initial parameter allocation strategy (\ding{193}).

During the online phase, the hardware configuration and batched inputs (\ding{194}) are provided to the ParaSpec Planner (\secref{sec:paraspec}), which, based on input length and characteristics, computes a fine-grained pipeline execution plan, including the batch sizes for target model in prefill and decoding phase, batch size for draft model, and the number of candidate tokens to generate (\ding{195}).

The scheduling results from both offline and online phases  (\ding{196},\ding{197}) collectively determine the Interleaved Batch Pipeline (\secref{sec:pipeline}). The pipeline consists of three main threads: GPU computation, CPU computation, and GPU-CPU I/O.  Parameter residency within GPU and CPU memory dynamically adapts in response to the progression of the I/O thread (\ding{198}).

\section{Method}

\subsection{Interleaved Batch Pipeline}
\label{sec:pipeline}
Motivated by the differing computational demands of the prefill and decoding phases in LLM inference, we introduce the Interleaved Batch Pipeline—a phase-specific pipeline design. During the prefill stage, the target model computation dominates the GPU runtime, and we perform additional memory management before completion. In the decoding stage, we embed speculative decoding into the pipeline by finely interleaving the computations of the two models to fully utilize the GPU cores.
 %While the prefill phase fully utilizes the GPU due to its high computational overhead, the decoding phase is comparatively lightweight, resulting in underutilized GPU resources. This idle capacity is effectively leveraged by embedding a draft model to enhance throughput.

\subsubsection{Prefill Phase}
% The pipeline design for the prefill phase primarily follows the "zig-zag"  strategy proposed by FlexGen \cite{Flexgen}, where multiple micro-batches are processed consecutively after loading a layer's parameters onto the GPU. After the prefill phase, we offload selected model parameters and the entire KV cache to CPU DRAM to free up GPU VRAM for draft model in speculative decoding.
%Based on prior studies \cite{Flexgen, neo, moe-lightning} and our experiments, we conclude that during the prefill phase, neither Attention nor FFN computations can be offloaded to the CPU. Consequently, all computations during prefill must be executed on the GPU. 
% Shortly, the prefill phase primarily involves two threads: one for parameter IO and one for GPU computation. We overlap them as much as possible.
While our pipeline design for the prefill phase is inspired by the "zig-zag" strategy proposed by FlexGen \cite{Flexgen}, we extend this approach by tailoring the micro-batch scheduling and parameter management to better support speculative decoding. To minimize the GPU memory footprint of the target model during the offloading stage, at the end of the prefill phase, we offload partial model parameters and the entire KV cache to CPU memory. 
% This allows the “low-yield” GPU memory in the decoding phase to be effectively utilized by the draft model.

\subsubsection{Decoding Phase}
\begin{figure}
    \centering
    \includegraphics[width=\linewidth]{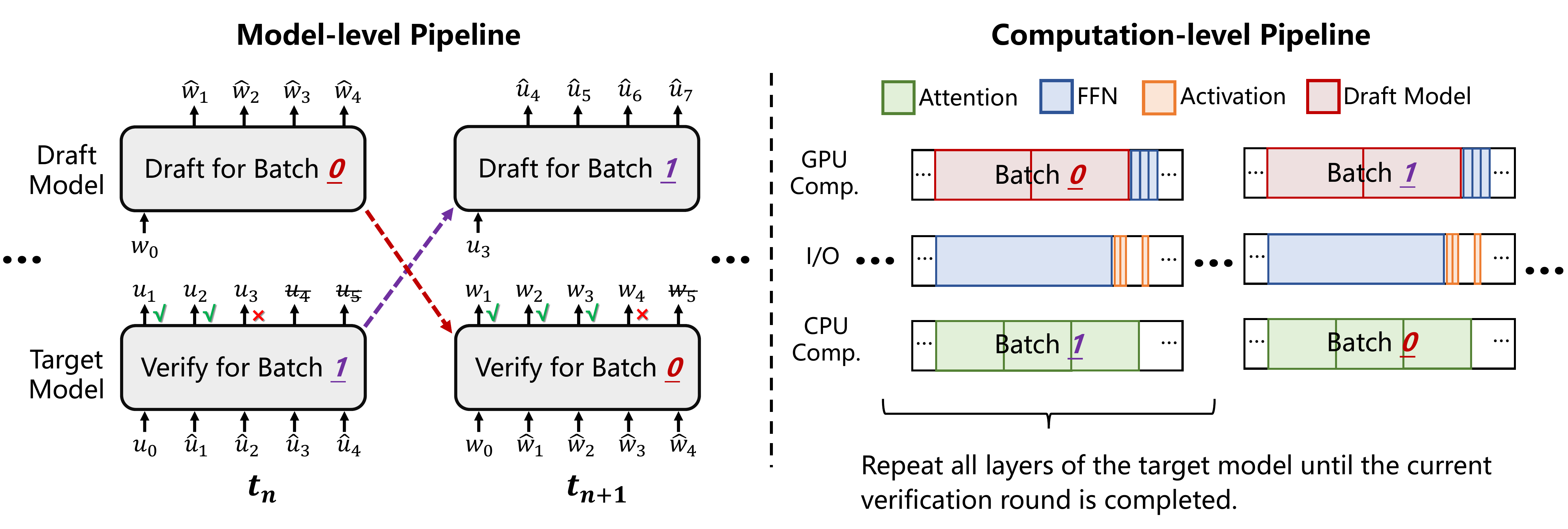}
    \caption{Schematic of the decoding pipeline. At model-level, while the target model validates Batch 1, the draft model concurrently generates tokens for Batch 0 (in time slot $t_n$); the two models then alternate batches (in time slot $t_{n+1}$). At computation-level, the target model's parameters are computed layer by layer. GPU, IO, and CPU are orchestrated to perform distinct, parallelized tasks. }
    \vspace{-12pt}
    \label{fig:pipeline}
\end{figure}

% The key difference between the decoding and prefill phases is that the computational load of LLM inference decreases, while I/O bottlenecks become more pronounced. To minimize data transfer overhead, the draft model for speculative decoding is kept resident in GPU memory during decoding, while attention computations of target model are offloaded to the CPU to avoid KV cache movement.
% (Since the KV cache for input tokens has already been stored during the prefill phase, the subsequent attention workload is significantly reduced and can be efficiently handled by CPU.)
During the decoding phase, we build upon the original offloading framework by repurposing the low-yield GPU memory to store the draft model and leveraging GPU idleness to execute it. To enable this, we redesign the entire pipeline at both the model and computation levels. The decoding phase pipeline is illustrated in \figref{fig:pipeline}. In summary, our model-level design transforms conventional single-batch speculative decoding into a dual-batch interleaved scheme to facilitate parallel execution. At computation-level, the draft model’s workload is finely interleaved into the GPU idle periods between the layer-wise computations of the target model.

% From model-level, the decoding phase involves two batches being processed alternately by the target and draft models. Conventional speculative decoding employs a single-batch Draft-then-Verify paradigm, which requires the target model to wait for the draft model to complete its computation, thereby inhibiting parallelism. As a result, merely incorporating single-batch speculative decoding into the offloading framework exacerbates both I/O and computational overhead. To address this, we propose a dual-batch rotation design. As illustrated in the left side of \figref{fig:pipeline}, while the target model verifies Batch 1, the draft model generates a speculative output for Batch 0 ($\hat{w}_1$, $\hat{w}_2$, $\hat{w}_3$, $\hat{w}_4$). Upon completing their respective tasks, the models swap batches: the target model validates the draft output of Batch 0 (${w}_1$, ${w}_2$, ${w}_3$, ${w}_4$), while the draft model begins generating the draft for Batch 1. The two batches are interleaved and processed alternately until the required number of tokens is generated. This batch interleaving primarily enables parallel execution of drafting and verification.
At model-level, the decoding phase involves two batches being processed alternately by the target and draft model. Conventional speculative decoding adopts a single-batch Draft-then-Verify paradigm, as the computations of the draft and target models must be executed sequentially. Consequently, the GPU resources left idle during the target model’s verification stage remain unused, while the system suffers from additional overhead caused by switching between models.

To overcome these limitations, we propose a dual-batch interleaved design that enables true model-level parallelism. As shown on the left side of \figref{fig:pipeline}, in time slot $t_n$, while the target model verifies Batch 1, the draft model concurrently generates speculative tokens for Batch 0 ($\hat{w}_1$, $\hat{w}_2$, $\hat{w}_3$, $\hat{w}_4$). Once both tasks are completed, the roles switch. In time slot $t_{n+1}$, the target model validates Batch 0 (${w}_1$, ${w}_2$, ${w}_3$, ${w}_4$), and the draft model proceeds with Batch 1. This alternating batch rotation continues until the generation is complete.
This batch interleaving primarily enables the parallel execution of drafting and verification.

At computation-level, the decoding pipeline involves coordination among three threads: GPU computation, CPU computation, and I/O. The right side of \figref{fig:pipeline} provides an illustrative example. Each layer of the target model is fine-grainedly partitioned into attention and FFN components. For clarity, lightweight components such as normalization layers are omitted from the figure, as their parameter size and computational cost are negligible. As illustrated, for the majority of the inference time, the CPU performs attention computation (current batch), the system transfers same layer's FFN parameters from CPU memory to GPU memory (current batch), and the GPU executes draft model computation (another batch), all in parallel. Upon completion of attention computation on the CPU, the intermediate activations are transferred to the GPU. Finally, once both the parameters and activations are available on the GPU, the remaining computations are quickly completed. After completing all layers of the target model (the end of a round of parallel drafting and verification at model-level), the target and draft models exchange their current batches. 
% This leads to a high degree of overlap among the three threads, resulting in a well-synchronized and efficient pipeline.

Details such as where model parameters are computed, how I/O is handled, and batch configurations are determined in the \secref{sec:placement} and \secref{sec:paraspec}. 

\subsection{Adaptive Tensor Placement}
\label{sec:placement}
% \methodname is built upon a heterogeneous memory hierarchy encompassing GPU VRAM, CPU DRAM, and even disk storage, where we further optimize the placement of target and draft model parameters to maximal resource utilization. 
% In contrast to prior work that focuses solely on parameter placement for a single model, \methodname introduces a novel design to accommodate speculative decoding by jointly managing the parameter placement of both the target and draft models across GPU memory, CPU memory, and disk tiers. 
% In contrast to prior work that focuses solely on parameter placement for a single model, 
\methodname introduces a novel design for heterogeneous models (target and draft models), by jointly managing the parameter placement across GPU memory, CPU memory, and disk tiers, thus enabling efficient speculative decoding.
Adaptive Tensor Placement strategy intelligently assigns tensors to different memory tiers based on real-time resource availability and the current computational task, optimizing memory utilization and mitigating I/O bottlenecks.

We establish tensor prioritization hierarchically by sub-layer, categorizing based on both functional type (attention, KV cache, FFN) and computational phase. Tensors required by the current and next layers of the target model are assigned the highest priority and are preferentially placed in GPU memory. Draft model and its cache are also treated as high-priority and retained in GPU memory during decoding phase. If GPU memory capacity permits, additional parameters are pinned to further reduce I/O overhead. Remaining tensors are offloaded to CPU memory with moderate priority, leveraging its high bandwidth and low latency, as well as its ability to support certain computations. If CPU memory is exhausted, parameters are further offloaded to disk. When CPU memory is sufficient, $pin\_memory()$ is employed to accelerate GPU-CPU data transfer.  A dynamic memory management mechanism is employed to avoid cross-tier memory swaps, ensuring that only CPU memory interfaces with both GPU memory and disk.

The core of the dynamic memory management mechanism is prefetching, which overlaps I/O with computation. For example, while computing attention of layer $i$, GPU memory preloads FFN of the same layer from CPU memory, and concurrently, CPU memory prefetches the parameters of layer $i+1$ from disk. Dedicated placeholders are reserved in GPU \& CPU memory for prefetched tensors. 
% It is worth noting that this work primarily targets scenarios with sufficient DRAM, and provides limited support for management mechanisms after offloading to disk.

\subsection{ParaSpec Planner}
\label{sec:paraspec}
Interleaved Batch Pipeline section (\secref{sec:pipeline}) outlines our pipeline strategy; however, key parameters—such as the batch sizes of target model during prefill and decoding phase, batch size of draft model,  generated draft token number, require careful tuning. To address this, we propose ParaSpec Planner, a parameter specialization module that selects optimal configurations for a given input.

\textbf{Planning Goal.} ParaSpec Planner aims to maximize model inference throughput on a given hardware configuration. Throughput is determined by two factors: the total number of tokens generated per batch inference, denoted as $\tilde{N}_{generated}$, and the corresponding generation latency, $T_{generation}$. On consumer-grade hardware, the primary system constraints lie in GPU memory capacity. Therefore, we formulate the problem as a constrained optimization task as follows:
\begin{align}
    \max ~ throughput &= \max~\frac{\tilde{N}_{generated}}{T_{generation}}
    \nonumber\\
    s.t.~~gpu~peak~memory~&\leq~gpu~mem~capacity 
\end{align}
\vspace{-1em}

\textbf{Generated Tokens.} The total number of generated tokens, $\tilde{N}_{{generated}}$, is the sum of tokens $\tilde{n}_{{generated}}$ produced over $n_{iter}$ iterations for a batch of size $bs$. However, in our system, speculative decoding introduces randomness, causing the number of tokens generated per input in each iteration to become a random variable. We use the expected value to represent the average number of tokens that pass verification in each iteration.
\begin{align}
    \tilde{N}_{generated}=\sum_{bs}\sum_{n_{iter}}\tilde{n}_{generated} = bs\times n_{iter}\times\mathbb{E}[n_{generated}]
\end{align}
\vspace{-1em}

\textbf{Inference Latency.} The inference latency $T_{generation}$ is determined by the degree of parallelism in the inference pipeline. Due to architectural differences between the prefill and decoding phases in \methodname, their latencies must be modeled separately, $T_{generation} = T_{prefill}+T_{decoding}$. Since computation in the prefill phase is primarily GPU-bound, its execution time is independent of batch size and instead depends on the number of computation steps required. 
\begin{align}
    T_{prefill}=\left\lceil\frac{bs}{bs_{prefill}}\right\rceil\times T_{target,prefill}^{GPU}
\end{align}
During the decoding phase, \methodname performs two primary tasks: draft generation for one batch and verification for another. The overall latency is determined by the longer of the two.
\begin{align}
    T_{decoding}=\max(T_{target,decoding},~~T_{draft})
\end{align}

\textbf{Memory Constraints.} GPU memory constraints can likewise be decomposed into those for the prefill and decoding phase. In each phase, the combined memory footprint of model parameters, intermediate activations, and KV cache must not exceed the available GPU memory. In the prefill phase, GPU memory consumption is primarily composed of two parts: the parameter size of the target model, and the KV cache required. 
\begin{align}
    V_{prefill}=V_{target,prefill}+V_{target,KVcache}
\end{align}
Similarly, in decoding, GPU memory usage consists of the main model parameters, the draft model parameters, and the KV cache used by the draft model.
\begin{align}
    V_{decoding}=V_{target,FFN}+V_{draft}+V_{draft,KVcache}
\end{align}

More detailed derivations, please refer to the \appendixref{Appendix:planner}. Before using the ParaSpec Planner, a profiling program must be run on the target hardware to collect performance characteristics. However, due to the challenges of hardware measurement, OS-induced variability, and the uncertainty in draft token validity introduced by speculative decoding, such measurements may not fully reflect the actual behavior of \methodname during execution. Consequently, while ParaSpec Planner can produce high-quality parameter configurations, further fine-tuning may still be required to achieve optimal performance. 

\section{Evaluation}

\begin{table}[ht]
    \centering
    \begin{minipage}{0.4\linewidth}
        \small
        \setlength{\tabcolsep}{3pt} % 减小列间距
        \caption{Hardware Configurations.}
        \label{tab:hardware}
        \begin{tabular}{>{\centering\arraybackslash}m{1cm} 
                        >{\centering\arraybackslash}m{2cm} 
                        >{\centering\arraybackslash}m{2cm}} 
        \toprule
         & \textbf{Env \#1} & \textbf{Env \#2} \\
        \midrule
        GPU & RTX 4090 & RTX 4090 \\
        VRAM & 24G & 24G \\
        PCIe & Gen3 x 16 & Gen4 x 16 \\
        CPU & i9-10980XE & EPYC 7542 \\
        DRAM & 256G & 448G \\
        \bottomrule
        \end{tabular}
    \end{minipage}%
    \hfill
    \begin{minipage}{0.56\linewidth}
        \small
        \setlength{\tabcolsep}{3pt} % 同样减小列间距
        \caption{Dataset Configurations.}
        \label{tab:dataset}
        \begin{tabular}{>{\centering\arraybackslash}m{0.8cm} 
                        >{\centering\arraybackslash}m{1.5cm} 
                        >{\centering\arraybackslash}m{1.5cm}
                        >{\centering\arraybackslash}m{1.5cm}
                        >{\centering\arraybackslash}m{1.5cm}} 
        \toprule
         & \textbf{HumanEval} \cite{humaneval} & \textbf{C-Eval} \cite{ceval} & \textbf{SummEval} \cite{summeval}  & \textbf{SAMSum} \cite{samsum} \\
        \midrule
        $S_{avg}$ & 157.54 & 165.46 & 503.02 & 168.10\\
        $S_{max}$ & 437    & 483    & 783    & 1144\\
        $S_{std}$ & 72.46  & 103.18 & 138.68 & 120.53\\
        % $S_{min}$ & 32     &115   &626     \\
        Task      & Coding & Exam     &  \multicolumn{2}{c}{Summarization}        \\
        \bottomrule
        \end{tabular}
    \end{minipage}
\end{table}

\subsection{Experimental Setup}
\label{sec:expsetup}
\textbf{Implementation.} We implement \methodname on top of Hugging Face Transformers v4.47.1\cite{transformers}. We implement pipeline using multiprocessing  with shared memory for inter-process vector communication. More details is provided in \appendixref{Appendix: implementation}.

\textbf{Models.} We evaluate \methodname using two popular and open-source models: Mixtral-8x7B \cite{Mixtral8x7b} and Mixtral-8x22B \cite{Mixtral8x22b}. They have 46.7B and 141B parameters in bfloat16 precision. The draft model for speculative decoding is Mistral-7B \cite{mistral7b}. Although not evaluated, \methodname can support other models compatible with Transformers' model classes.

\textbf{Hardware.} We evaluate \methodname in two different evironments, as shown in \tabref{tab:hardware}. Env \#2 refers to a cloud-based server. 
% In both environment, ample CPU memory resources are available; therefore, read/write speeds of the disk need not be considered. Env \#2 refers to a cloud-based server.

\textbf{Datasets.} We use most common LLM benchmarks with different prompt length distributions and tasks, as shown in \tabref{tab:dataset}. HumanEval dataset \cite{humaneval} released by OpenAI includes 164 programming problems; C-Eval dataset \cite{ceval} is a comprehensive Chinese evluation suite includes 13948 questions; SummEval dataset \cite{summeval} inlcudes 100 news article from the CNN/DailyMail; SAMSum dataset \cite{samsum} contains about 16k messenger-like conversations with summaries.

\textbf{Baselines.} We compare against 4 baseline systems, all designed to address GPU memory limitations. 
\begin{itemize}[leftmargin=2em, topsep=-2pt, parsep=0pt]
    \item Hugging Face Accelerate \cite{accelerate} supports offloading weights of some layers based on the device map. We use its version 1.5.2. Hereinafter referred to as Accelerate.
    \item DeepSpeed Zero-Inference \cite{deepspeed} supports offloading the whole weights to the CPU or disk. We use its version 0.16.1. Hereinafter referred to as DeepSpeed.
    \item FlexGen \cite{Flexgen} employs a "zig-zag" inference schedule to increase throughput.
    \item Fiddler \cite{fiddler} strategically utilizes both CPU and GPU resources for MoE model inference.
\end{itemize}

Additionally, except for FlexGen, all other baselines natively support  Mixtral. We adapted FlexGen to Mixtral while adhering to its original offloading strategy. 

\textbf{Metrics.} Throughput (token/s) is calculated as the number of tokens generated divided by total genration time (prefill time + decoding time).

\subsection{End-to-end Throughput}
\begin{figure}
    \centering
    \includegraphics[width=\linewidth]{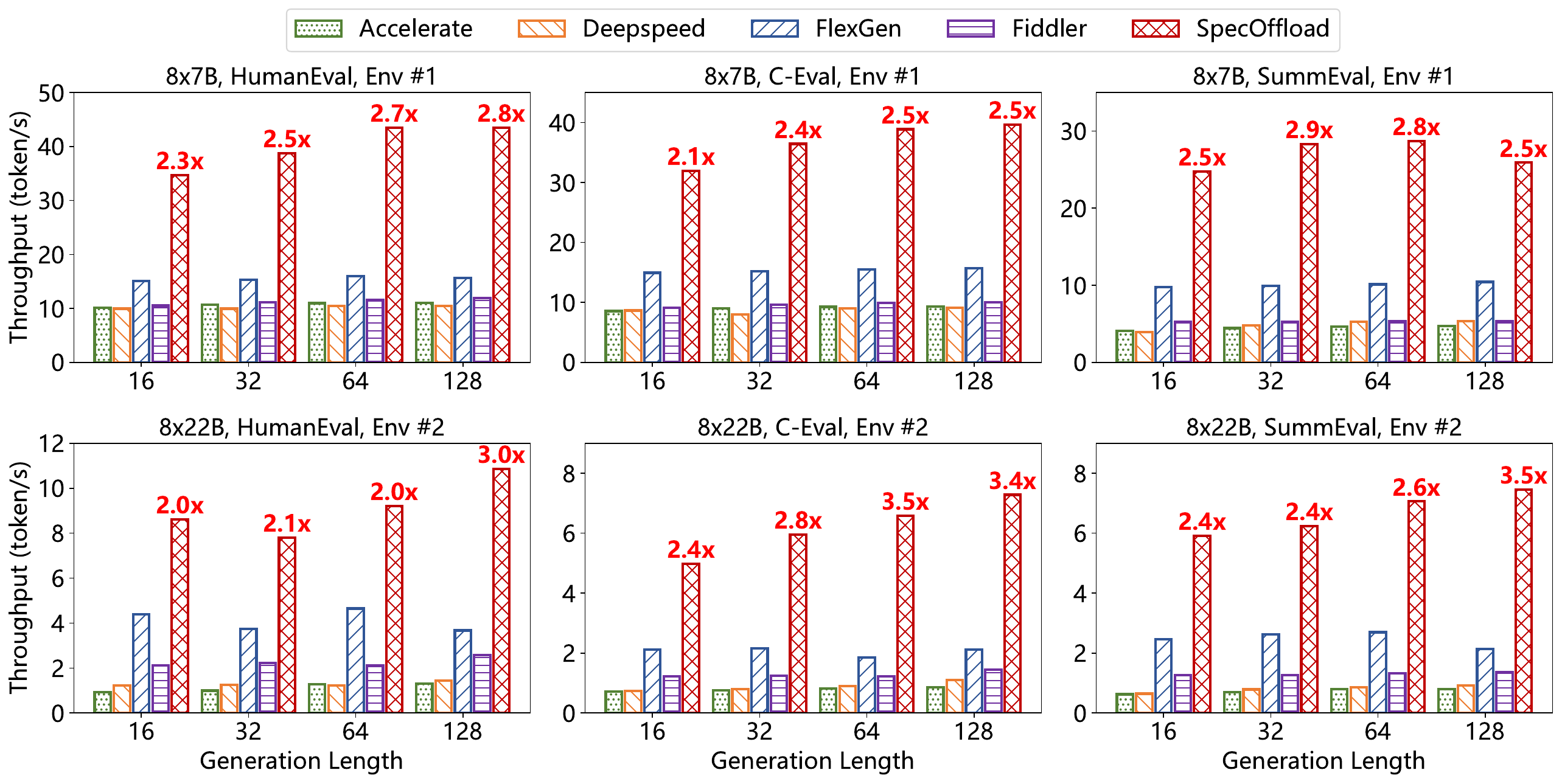}
    \caption{End-to-end comparison between \methodname and baselines in different scenarios.}
    \vspace{-10pt}
    \label{fig:overall_performance}
\end{figure}
\figref{fig:overall_performance} illustrates the end-to-end throughput of five approaches across two environments and three datasets. In Env \#1, using the Mixtral 8×7B model, \methodname attains an average speedup of 2.53× over the strongest baseline, FlexGen; in Env \#2, with the Mixtral 8×22B model, it achieves an average speedup of 2.54×. The superior throughput of \methodname is attributed to the embedding of speculative decoding into the offloading pipeline. Empirical results demonstrate that computing and storing the draft model on the 
GPU memory confers greater performance gains than offloading either the target model’s parameters or its cache.

% Overall, throughput increases with the number of generated tokens for all methods, as longer decoding phases amortize the cost of the expensive prefill phase. Unlike other baselines, our method incorporates speculative decoding, whose effectiveness may vary with dataset difficulty, thus impacting throughput. When the draft model correctly predicts more tokens, \methodname achieves higher throughput. The acceleration from speculative decoding is most pronounced on the HumanEval dataset.

% Theoretically, \methodname is expected to perform better with the 8×22B model; however, the observed improvements over the baseline are similar for both 8×7B and 8×22B. As analyzed in the Motivation section, larger models yield lower cost-effectiveness in terms of parameter and cache storage. Nevertheless, in Env \#2, faster I/O channels and weaker CPU performance narrow the performance gap between \methodname and the baseline.

The performance of our method on additional datasets is provided in the \appendixref{Appendix:througput}. More Details on the policy setups and effective batch sizes can be found in \appendixref{Appendix:Policy}.

\subsection{Effectiveness Analysis}
\begin{table}[htbp]
    \centering
    \caption{Detail Runtime breakdown (seconds)."P" and "D" denote Prefill and Decoding, respectively. Compute(G,T) and Compute(G,D) denote the GPU computation time for the target and draft models, respectively, while Compute(C) represents the target model’s computation time on the CPU. Cache(G→C) indicates the time to transfer the KV cache from GPU memory to CPU memory.
}
    \begin{tabular}{>{\centering\arraybackslash}m{1.08cm}
                >{\centering\arraybackslash}m{0.5cm}
                >{\centering\arraybackslash}m{0.6cm}
                >{\centering\arraybackslash}m{1.8cm}
                >{\centering\arraybackslash}m{1.8cm}
                >{\centering\arraybackslash}m{1.5cm}
                >{\centering\arraybackslash}m{1.1cm}
                >{\centering\arraybackslash}m{2cm}}
    \toprule
    \textbf{} & Phase & Total & Compute(G,T) & Compute(G,D) & Compute(C) &  Weight(R) & Cache(G→C)\\
    \midrule
    \multirow{2}{=}{
    \centering {8×7B, Env \#1}} & P & 183.28 & 79.62 & 0 & 0 & 123.48 & 39.05 \\
        & D & 569.21 & 35.34 & 489.02 & 531.23 & 236.2 & 0 \\
    \midrule
    \multirow{2}{=}{
    \centering {8×22B, Env \#2}} 
    & P & 280.42 & 42.22 & 0 & 0 & 166.45 & 91.06\\
    & D & 794.26 & 27.34 & 345.93 & 746.38 & 262.64 & 0\\
    \bottomrule
    \end{tabular}
    \label{tab:breakdown}
\end{table}

\vspace{10pt}

\begin{figure}[htbp]
    \centering
    \setlength{\abovecaptionskip}{0pt} %
    \setlength{\belowcaptionskip}{2pt}% caption和图之间的距离
    \begin{minipage}{0.49\textwidth}
        \centering
        \includegraphics[width=\textwidth]{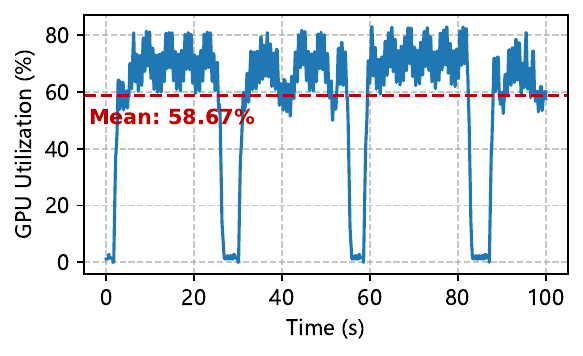} % 图片路径
        \caption{Decoding phase GPU core utilization.}
        \label{fig:gpu-util}
    \end{minipage}
    \hfill
    \begin{minipage}{0.49\textwidth}
        \centering
        \includegraphics[width=\textwidth]{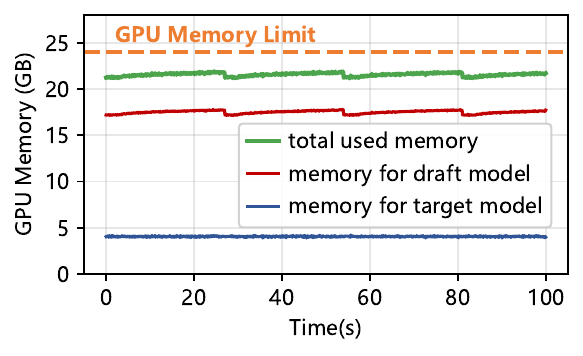} % 图片路径
        \caption{Decoding phase memory consumption.}
        \label{fig:gpu-memory}
    \end{minipage}
    % \begin{minipage}{0.5\textwidth}
    %     \centering
    %     \small
    %     \setlength{\tabcolsep}{3pt} % 减小列间距
    %     \captionof{table}{Runtime breakdown (seconds). "P" and "D" denote prefill and Decoding, respectively.}
    %     \label{tab:runtime}
    %     \begin{tabular}{>{\centering\arraybackslash}m{1.1cm}
    %             >{\centering\arraybackslash}m{0.9cm}
    %             >{\centering\arraybackslash}m{0.9cm}
    %             >{\centering\arraybackslash}m{0.9cm}
    %             >{\centering\arraybackslash}m{0.9cm}
    %             >{\centering\arraybackslash}m{0.9cm}}
    % \toprule
    % \textbf{} & \textbf{Phase} &\textbf{Total} & \textbf{GPU Comp} & \textbf{IO} & \textbf{CPU Comp} \\
    % \midrule
    % \multirow{2}{=}{
    % \centering {8×7B in Env \#1}} & P & 183.28 & 79.62  & 162.53 & 0\\
    %                           & D & 569.21 & 524.36 & 236.20 & 531.23 \\
    % \midrule
    % \multirow{2}{=}{\centering {8×22B in Env \#2}} & P & 280.42 & 42.22 & 257.51 & 0 \\
    %                                           & D   & 794.26 & 373.27 & 262.64 & 746.38 \\
    % \bottomrule
    % \end{tabular}
    % \end{minipage}
\end{figure}
\vspace{5pt}

We employ NVIDIA Nsight \cite{nsight} to monitor GPU core utilization and memory consumption during the decoding phase of Mixtral 8×7B in Env \#1 on the SummEval dataset. As depicted in \figref{fig:gpu-util}, the average core utilization reaches 58.67\%, attains 8.14×, 7.15×, 4.49×, and 8.24× higher than Accelerate, DeepSpeed, FlexGen, and Fiddler, respectively. The elevated core utilization can be attributed to the residency of the draft model in GPU memory, enabling continuous computation.

\figref{fig:gpu-memory} reveals a periodic pattern in the draft model’s GPU memory consumption. Each cycle lasts approximately 28 seconds, characterized by a gradual increase in memory usage followed by a sharp drop and a 2-second idle window. This aligns with the behavior observed in \figref{fig:gpu-util}, where the draft model performs computation for 26 seconds and remains idle for 2 seconds awaiting the next batch. More detailed GPU memory use in \appendixref{Appendix:vram}.
%\figref{fig:gpu-util} clearly shows a periodic pattern in GPU utilization, with brief idle periods occurring every $27\sim28$ seconds, during which GPU usage drops near zero. These idle periods indicate that the draft model has completed generating the next draft token, while the large model's parameters are still being loaded via IO. Although allowing the draft model to predict more tokens can increase GPU utilization, scheduling results show that it does not improve throughput.

We show the runtime breakdown of Mixtral 8x7B in Env \#1 and 8x22B in Env \#2 in \tabref{tab:breakdown} on SummEval dataset. We disable overlapping and profile the time used for GPU, I/O, and CPU. Results show that our method effectively overlaps compute and I/O. 
% Using 8×7B decoding as a case study, GPU computation, I/O, and CPU computation account for 92.1\%, 41.5\%, and 93.3\% of the total time, respectively. The results for 8×22B are similar; however, during the decoding phase, performance is more likely bottlenecked by CPU computation, as its latency significantly exceeds that of GPU computation and I/O. 

\subsection{Ablation Study}
\begin{table}[ht]
    \centering
    \setlength{\tabcolsep}{3pt} % 同样减小列间距
    \caption{Ablation study of proposed techniques on SummEval dataset. The gray tuple denotes a policy (prefill batch size, decoding batch size, draft batch size, draft max new tokens).}
    \label{tab:ablation}
    \begin{tabular}{>{\centering\arraybackslash}m{0.8cm} 
                    >{\centering\arraybackslash}m{3.0cm} 
                    >{\centering\arraybackslash}m{3.0cm}
                    >{\centering\arraybackslash}m{3.2cm} 
                    >{\centering\arraybackslash}m{3cm}}
    \toprule
     & All optimizations & No policy search & Serial SD & No SD \\
    \midrule
    8x7B & 24.743 \textcolor{gray}{(80, 192, 8, 8)}& 15.624 \textcolor{gray}{(50, 256, 5, 2)} & 17.048 \textcolor{gray}{(80, 192, 40, 8)}& 12.369 \textcolor{gray}{(80, 256, x, x)} \\
    8x22B & 5.911 \textcolor{gray}{(16, 64, 8, 8)}  & 3.486 \textcolor{gray}{(16, 32, 6, 6)}& 4.146 \textcolor{gray}{(16, 64, 32, 8)} & 1.698 \textcolor{gray}{(16, 80, x, x)} \\
    \bottomrule
    \end{tabular}
\end{table}
\vspace{8pt}

We then isolate the improvement brought by each individual technique. \tabref{tab:ablation} lists the throughput \methodname can achieve if disabling one technique at a time. On Mixtral 8x7B, we choose target model prefill batch 80, decoding batch size of 192. And due to the design of rotating batches, the total batch size is 192 $\times$ 2 = 384. Draft model batch of 8 and generates 8 tokens per iteration. "No policy search" illustrates the performance of a random strategy, showing the importance of a good policy. %More Details on the policy setups and effective batch sizes can be found in \appendixref{Appendix:Policy}. 
We also decoupled speculative decoding from the tightly integrated pipeline into a loosely coupled serial mode. Results show that embedding speculative decoding into the pipeline is beneficial, as the naive combination of serial speculative decoding and offloading incurs additional I/O overhead from the draft model and its KV cache. Removing speculative decoding highlights its substantial performance gains. Additional ablation studies are provided in \appendixref{Appendix:ablation}.

\subsection{Load to disk}
\begin{figure}[htbp]
    \centering
    \includegraphics[width=\linewidth]{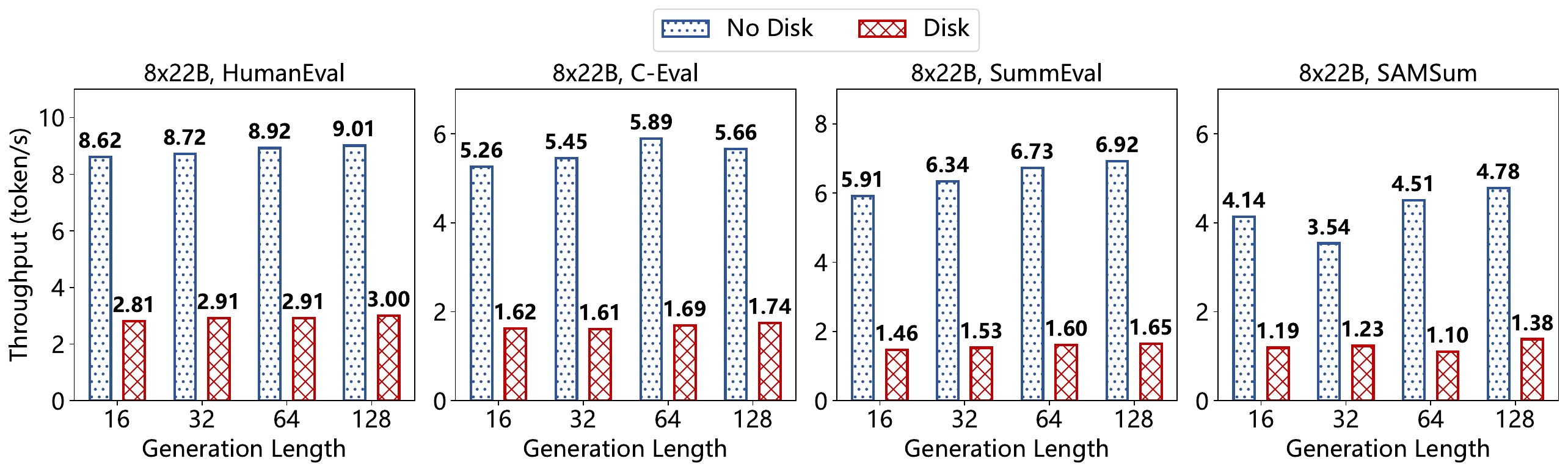}
    \caption{Throughput comparison of Mixtral 8×22B with and without disk offloading. No Disk corresponds to execution in Env \#2, which has sufficient CPU memory, while Disk corresponds to execution in Env \#1, where GPU memory is limited.}
    \label{fig:appedix_disk}
\end{figure}

We further conducted experiments in Env \#1 by extending the offloading to disk. The disk read and write speeds are 3.5GB/s and 1.7GB/s, respectively. Env \#1 offers a memory capacity of 256GB, which is insufficient to accommodate the complete Mixtral 8×22B model (141B parameters). As shown in \figref{fig:appedix_disk}, under these memory-limited conditions, load to disk \methodname attains 29.3\% of its original throughput.

\section{Conclusion}

We identify two key inefficiencies in existing offloading frameworks for LLM inference: underutilization of GPU cores and marginal utility of GPU memory. To address these, we propose \methodname, which embeds speculative decoding into offloading with virtually zero overhead by leveraging idle GPU time and "low-yield" GPU memory. Experiments show up to 2.54× throughput gains over the best baseline, demonstrating the effectiveness of our approach for high-throughput LLM inference on resource-constrained devices.

\newpage
\bibliographystyle{unsrt}  % 或者 alpha, ieee, apalike 等
\bibliography{reference}   % 不带 .bib 后缀

%%%%%%%%%%%%%%%%%%%%%%%%%%%%%%%%%%%%%%%%%%%%%%%%%%%%%%%%%%%%

\newpage
\appendix
\section{Technical Appendices and Supplementary Material}

\subsection{ParaSpec Planner}
\label{Appendix:planner}
\textbf{Planning Goal.} 

ParaSpec Planner aims to maximize model inference throughput on a given hardware configuration. Throughput is determined by two factors: the total number of tokens generated per batch inference, denoted as $\tilde{N}_{generated}$, and the corresponding generation latency, $T_{generation}$. On consumer-grade hardware, the primary system constraints lie in GPU memory capacity. Therefore, we formulate the problem as a constrained optimization task as follows:
\begin{align}
    \max ~ throughput &= \max\frac{\tilde{N}_{generated}}{T_{generation}}
    \nonumber\\
    s.t.~~gpu~peak~memory~&\leq~gpu~mem~capacity 
    \label{eq:total}
\end{align}

\textbf{Generated Tokens.} 

The total number of generated tokens, $\tilde{N}_{{generated}}$, is the sum of tokens $\tilde{n}_{{generated}}$ produced over $n_{{iter}}$ iterations for a batch of size $bs$ . In conventional decoding without speculation, each input generates exactly one token per iteration, making $\tilde{n}_{{generated}}$ a constant. However, in our system, speculative decoding introduces randomness, causing the number of tokens generated per input in each iteration to become a random variable. As a result, $\tilde{n}_{{generated}}$ cannot be expressed deterministically and is instead characterized by its expected value. Let $\tilde{n}_{{generated}}$ denote the actual number of tokens generated for a single input, then $\tilde{n}_{{generated}} = \mathbb{E}[n_{{generated}}]$, as shown in \equref{eq:generate_E}.

\begin{align}
    \tilde{N}_{generated} &=\sum_{bs}\sum_{n_{iter}}\tilde{n}_{generated} \\
    &=bs\times n_{iter}\times\mathbb{E}[n_{generated}]
    \label{eq:generate_E}
\end{align}

To characterize the distribution of $\tilde{n}_{\text{generated}}$, we model the speculative decoding process. In each iteration, the draft model generates a candidate sequence of $n_{\text{cand}}$ tokens, which is then verified by the target model. The target model returns the longest correct prefix of the candidate sequence and subsequently generates one additional correct token. The number of tokens correctly predicted by the draft model ranges from $0$ to $n_{\text{cand}}$, so $\tilde{n}_{\text{generated}}$ follows a distribution over the set $\{1, \dots, n_{\text{cand}} + 1\}$.

We assume that the probability of the draft model correctly predicting a single token is $p$, and that these predictions are independent across positions. Under this assumption, the probability that the main model accepts exactly $k$ tokens is given by the probability that the first $k-1$ tokens are correct and the $k$th is incorrect, as shown in \equref{eq:n_generate_wrong}. If $k = n_{\text{cand}} + 1$, it corresponds to the draft model correctly predicting the entire candidate sequence, this probability distribution is formalized in in \equref{eq:n_generate_True}. 

\begin{align}
    \mathbb{P}[n_{generated}=k]=p^{k-1}\cdot(1-p_{cand}),\quad k=1,\cdots,n_{cand}
    \label{eq:n_generate_wrong}
\end{align}
\begin{align}
    \mathbb{P}[n_{generated}=k]=p^{k-1},\quad k=n_{cand}+1
    \label{eq:n_generate_True}
\end{align}

The expected value $\mathbb{E}[n_{\text{generated}}]$ is derived in \equref{eq:ge_d}. Thus, the total number of tokens generated by the model, $\tilde{N}_{\text{generated}}$, is expressed as a function of $bs$, $n_{\text{iter}}$, $n_{\text{cand}}$, and $p$.

\begin{align}
    \mathbb{E}[n_{generated}] &=\sum_{k=1}^{n_{cand}+1}k\cdot\mathbb{P}[n_{generated}=k]
    \nonumber\\
    &=\frac{1}{1-p}[n_{cand}p^{n_{cand}+2}-(n_{cand}+1)p^{n_{cand}+1}+1]
    \label{eq:ge_d}
\end{align}

\textbf{Inference Latency.}

The inference latency $T_{\text{generation}}$ is determined by the degree of parallelism in the inference pipeline. As SpecOffload exhibits distinct behaviors in the Prefill and decoding stages, their latencies must be computed separately, in \equref{eq:t=p+d}.
\begin{align}
    T_{generation}=T_{prefill}+T_{decoding}
    \label{eq:t=p+d}
\end{align}

In the Prefill stage, loading the full KV cache for all $bs$ inputs would exceed GPU memory capacity. Therefore, SpecOffload partitions the batch into small Prefill batch $bs_{{prefill}}$ . Since computation in the Prefill stage is primarily GPU-bound, its latency is independent of the Prefill batch size and instead determined by the number of iterations required, as formalized in \equref{eq:t_Prefill}.

\begin{align}
    T_{prefill}=\left\lceil\frac{bs}{bs_{prefill}}\right\rceil\times T_{target,prefill}^{GPU}
    \label{eq:t_Prefill}
\end{align}

In each iteration, the processing time per $bs_{\text{Prefill}}$ is primarily determined by parameter I/O ($T^{C2G}_{para}$) and computation ($T^{GPU}_{target, comp}$), with I/O time significantly exceeding computation time in the offloading scenario, as shown in \equref{T_target,Prefill}.

\begin{align}
    T_{target,prefill}^{GPU} = T^{C2G}_{para}+T^{GPU}_{target, comp} \approx T^{C2G}_{para}
    \label{T_target,Prefill}
\end{align}

In the decoding stage, \methodname performs two primary tasks in parallel: draft generation for one batch and verification for another. The overall latency is thus determined by the slower of the two tasks in \equref{eq:t_decoding}.

\begin{align}
    T_{decoding}=\max(T_{target,decoding},~~T_{draft})
    \label{eq:t_decoding}
\end{align}

The draft generation task incurs a latency equal to the time required to execute the draft model inference entirely on the GPU. Similarly, due to memory constraints, the draft model must also divide each batch into smaller sub-batches $bs_{draft}$ for generation, $T_{draft}^{GPU}$ is the time for one-batch generation. Each generation step can be further decomposed into Prefill and decoding stages, as shown in \equref{eq:t_draft}.

\begin{align}
    T_{draft} &= \left\lceil\frac{bs}{bs_{draft}}\right\rceil\times
    T_{draft}^{GPU}
    \nonumber\\
    &=\left\lceil\frac{bs}{bs_{draft}}\right\rceil\times[T_{draft,prefill}^{GPU}+(n_{cand}-1)T_{draft,decoding}^{GPU}]
    \label{eq:t_draft}
\end{align}

For the verification task, based on \methodname’s pipeline design, each decoder layer's FFN computation depends on both the output of the Attention module and the loading of FFN parameters. Attention computation and FFN loading are executed in parallel threads, with the completion time determined by their maximum. Subsequently, the GPU performs FFN computation, which is significantly faster than parameter loading. Therefore, its latency can be expressed as in \equref{eq:t_target}. 
\begin{align}
    T_{target, decoding} &=n_{layer}\times[\max(T_{target,Attention}^{CPU},T_{target,FFN}^{C2G})+T_{target,FFN}^{GPU}] \nonumber\\
    &\approx n_{layer}\times[\max(T_{target,Attention}^{CPU},T_{target,FFN}^{C2G})]
    \label{eq:t_target}
\end{align}

Importantly, since the Attention module is offloaded to the CPU, its runtime becomes dependent on the sub-batch size, and is modeled accordingly in \equref{eq:t_target_mha}.

\begin{align}
    T_{target,Attention}^{CPU}=n_{cand}\times bs\times t_{target,Attention}^{CPU}
    \label{eq:t_target_mha}
\end{align}

\textbf{Memory constraint.}

The GPU memory capacity constraint can similarly be decomposed into separate constraints for the prefill and decode phases. In both phases, the total GPU memory consumption—including model parameters, intermediate activations, and KV cache—must not exceed the available GPU memory, which directly impacts the feasible batch size per inference.

In the prefill phase, GPU memory usage primarily consists of two components: the memory footprint of the main model parameters ($V_{\text{main}}$) and the KV cache required during prefill ($V_{m,\text{KVcache}}$). Since the KV cache for all $bs$ inputs would exceed GPU memory capacity, the batch is partitioned into sub-batches of size $bs_{\text{prefill}}$. As a result, the KV cache footprint in the prefill phase only accounts for $bs_{\text{prefill}}$ inputs, as formalized in \equref{eq:v_Prefill}.

\begin{align}
    V_{prefill}&=V_{target,prefill}+V_{target,KVcache} \nonumber\\ &=V_{target,prefill}+bs_{prefill}\times l_{input}\times v_{target,KVcache}
    \label{eq:v_Prefill}
\end{align}

Similarly, GPU memory usage in the decode phase consists of three components: the main model parameters loaded into GPU memory, the draft model parameters, and the KV cache used by the draft model. According to SpecOffload’s offloading strategy, only the MoE FFN parameters from the main model are loaded into GPU memory, whereas the draft model is fully resident in GPU memory. Therefore, the total GPU memory footprint during decoding is characterized in \equref{v_decoding1}. To satisfy the memory constraint, the batch of $bs$ inputs is partitioned into sub-batches of size $bs_{\text{assist}}$, as defined in \equref{v_decoding2}.

\begin{align}
    V_{decoding} &=V_{target,FFN}+V_{draft}+V_{draft,KVcache}
    \label{v_decoding1}
    \\
    &=V_{target,FFN}+V_{draft}+bs_{draft}\times(l_{input}+n_{generated})\times V_{draft,KVcache}
    \label{v_decoding2}
\end{align}

\newpage
\subsection{Implementation}
\label{Appendix: implementation}
\begin{figure}[htbp]
    \centering
    \includegraphics[width=\linewidth]{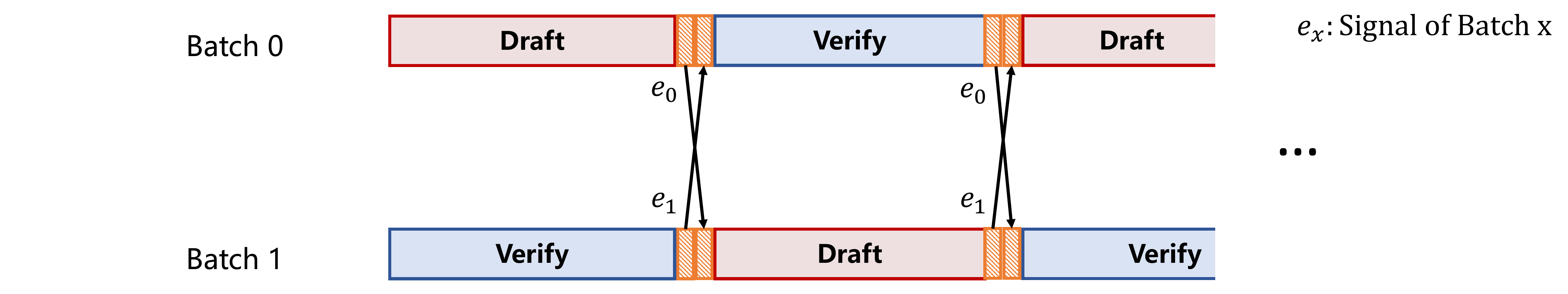}
    \caption{Implementation from the perspective of Interleaved batches.}
    \label{fig:appendix_macro_imp}
\end{figure}
\vspace{1em}

Our implementation is based on modifications to HuggingFace Transformers \cite{transformers}, version 4.47.1.

To implement the \methodname pipeline, we adopt a hybrid parallelism strategy that combines process-level and thread-level parallelism. As shown in \figref{fig:appendix_macro_imp}, the input sequence is split into two interleaved batches, which alternate between draft generation and large-model verification. Each batch is processed on a separate thread, with synchronization managed via inter-thread events. After completing its generation and verification task in each iteration, a thread signals its completion and waits for the other thread to do the same. The next iteration begins only after both threads have finished the current one. This design enables parallel execution of Batch 0's draft generation and Batch 1's verification. However, from the perspective of a single batch (Batch 0 or Batch 1), the draft and verify stages remain sequential.

In this design, the draft model performs full-sequence autoregressive inference on GPU during the generation stage, while the large model remains computing on CPU to avoid resource contention. Given that the draft model executes strictly in a sequential manner—processing only one batch at any time—parallel instantiation is unnecessary. To further minimize GPU memory consumption, only a single copy of the draft model is loaded and isolated within a dedicated auxiliary process. Communication between this draft model process and the main process hosting the large model is established via shared memory, enabling low-latency data transfer, while inter-process events are employed to enforce strict execution ordering and synchronization.

\begin{figure}[htbp]
    \centering
    \includegraphics[width=\linewidth]{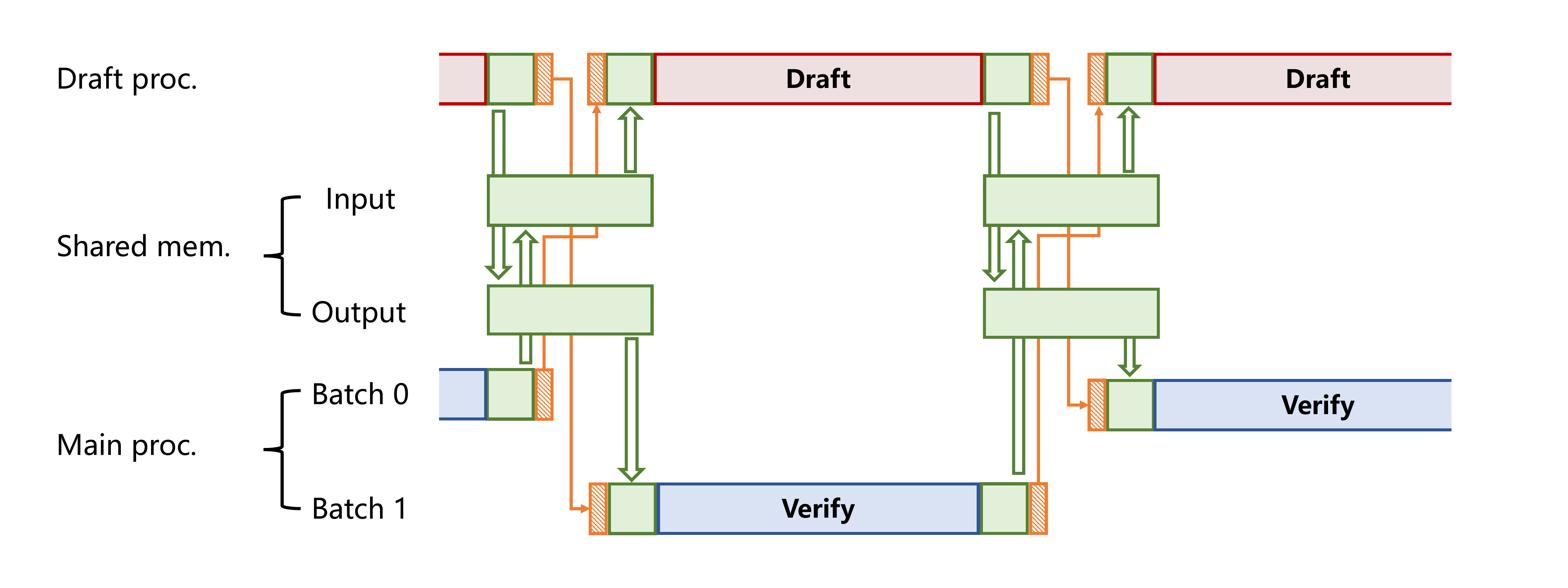}
    \caption{Inter-process communication diagram. Orange blocks represent the send/receive modules within each process, while green blocks indicate the inter-process communication modules. "Input" and "Output" are defined with respect to the Draft process.}
    \label{fig:appendix_micro_imp}
\end{figure}
\vspace{1em}
As shown in \figref{fig:appendix_micro_imp}, inter-process communication is centered around two shared memory regions, with tokens as the primary data exchanged. The Draft process writes generated draft tokens to the output shared memory for consumption by the Main process, and reads verified tokens from the input shared memory region. When a thread in the Main process reaches the verify stage and requires draft tokens, it waits for the draft model to signal availability, then reads from shared memory. Upon finishing computation, the draft model signals completion and readiness for the next task using event flags, ensuring proper synchronization and preventing data races or overlap between batches. This tightly coordinated mechanism enables efficient and orderly pipelined execution across model components, while keeping both memory footprint and runtime overhead to a minimum.

\subsection{Additional Experimental Results}

\subsubsection{End-to-end Throughput}
\label{Appendix:througput}
\begin{figure}[htbp]
    \centering
    \includegraphics[width=0.95\linewidth]{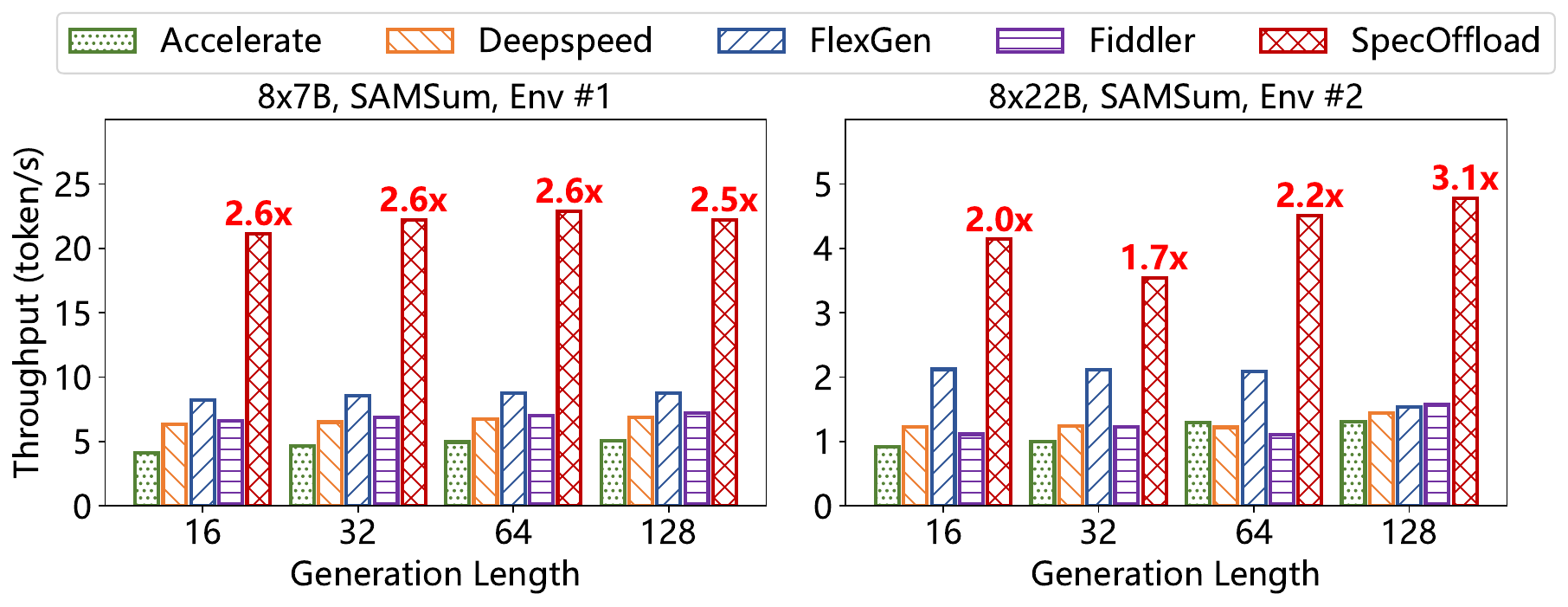}
    \caption{End-to-end comparison between \methodname and baselines on SAMSum.}
    \label{fig:appendix_throughput}
\end{figure}
\vspace{1em}

In addition to the datasets discussed in the main text, we further evaluated \methodname on the SAMSum dataset \cite{samsum}, which has an average input length of 168.1 and a maximum length of 1144. The SAMSum dataset contains about 16k messenger-like conversations with summaries. The results on SAMSum are consistent with those observed on the other datasets. As shown in \figref{fig:appendix_throughput}, \methodname achieves a throughput of 2.575× that of the best baseline on the Mixtral 8×7B model, and 2.25× on the Mixtral 8×22B model.

\subsubsection{Impact of Policy}
\label{Appendix:Policy}
We present detailed end-to-end throughput data, as shown in \tabref{tab:appendix_8x7B_policy_humaneval}, \tabref{tab:appendix_8x7B_policy_ceval}, \tabref{tab:appendix_8x7B_policy_summeval}, \tabref{tab:appendix_8x22B_policy_humaneval}, \tabref{tab:appendix_8x22B_policy_ceval}, and \tabref{tab:appendix_8x22B_policy_summeval}, to simulate different scenarios and analyze the impacts of policy on throughput by generating 16 tokens. Due to the large number of GPU hours required to complete all  (prefill batch size, decoding batch size, draft batch size, draft max new tokens) combinations using Mixtral 8×22B in Env \#2, 
we evaluated only a subset of possible configurations.

The prefill batch size is a tunable parameter for which the optimal value can be explicitly determined by the scheduling algorithm, as the Prefill stage does not involve speculative decoding and thus is free from probabilistic uncertainty. Under the experimental setup of \tabref{tab:appendix_8x7B_policy_summeval}, the optimal value is 80. For example, comparing entries 5 and 27 in \tabref{tab:appendix_8x7B_policy_summeval}—where all other parameters are held constant—the higher throughput of entry 27 is attributed to its more optimal Prefill batch size.

The decoding batch size and draft max new tokens jointly affect the verification latency of the target model. Since the target model's computation is offloaded to the CPU, the speculative decoding verification cannot achieve the same level of tight serialization as on the GPU. As a result, increasing the batch size or the number of new tokens leads to longer CPU computation time. As illustrated by entries 26–30 and 10, 20, 30, 40 in \tabref{tab:appendix_8x7B_policy_summeval}, neither a larger batch size nor a higher max new token value consistently yields better performance.

The decoding batch size, draft batch size, and draft max new tokens jointly impact the generation latency of the draft model. Due to GPU memory constraints, the draft batch size is typically limited to a small value. However, since all draft model computations are executed on the GPU, they are highly efficient. This allows the full decoding batch to be processed through a fine-grained, multi-round strategy. As long as the draft model's token generation time remains below the I/O-bound latency, it does not constitute a performance bottleneck.

The results in \tabref{tab:appendix_8x22B_policy_summeval} exhibit a similar pattern. These four parameters are tightly coupled and collectively determine the overall system throughput.
Given that our design introduces at least four tunable parameters, finding optimal settings through enumeration or heuristics alone is highly unlikely. This highlights the critical role of the Paraspec Planner in the overall system.
\begin{table}[htbp]
    \centering
    \setlength{\tabcolsep}{3pt} % 同样减小列间距
    \caption{Impact of policy on Mixtral 8x7B in Env \#1, HumanEval dataset.}
    \label{tab:appendix_8x7B_policy_humaneval}
    \begin{tabular}{>{\centering\arraybackslash}m{1cm} 
                    >{\centering\arraybackslash}m{2.5cm} 
                    >{\centering\arraybackslash}m{2.5cm}
                    >{\centering\arraybackslash}m{2.5cm}
                    >{\centering\arraybackslash}m{2.5cm}
                    >{\centering\arraybackslash}m{2cm}}
    \toprule
         No. & Prefill Batch Size & Decoding Batch Size & Draft batch size & Draft max new token & \textbf{Throughput (token/s)}\\
    \midrule
    1 & 80 & 200 & 10 & 8 & \textbf{32.821}\\
    2 & 80 & 160 & 6 & 1 & \textbf{15.869}\\
    3 & 80 & 160 & 6 & 2 & \textbf{20.964}\\
    4 & 80 & 160 & 6 & 4 & \textbf{28.914}\\
    5 & 80 & 160 & 6 & 6 & \textbf{33.711}\\
    6 & 80 & 160 & 6 & 8 & \textbf{33.690}\\
    7 & 80 & 160 & 8 & 1 & \textbf{15.834}\\
    8 & 80 & 160 & 8 & 2 & \textbf{20.940}\\
    9 & 80 & 160 & 8 & 4 & \textbf{29.267}\\
    10 & 80 & 160 & 8 & 6 & \textbf{32.520}\\
    11 & 80 & 160 & 8 & 8 & \textbf{32.776}\\
    12 & 80 & 160 & 10 & 1 & \textbf{15.835}\\
    13 & 80 & 160 & 10 & 2 & \textbf{21.120}\\
    14 & 80 & 160 & 10 & 4 & \textbf{29.499}\\
    15 & 80 & 160 & 10 & 6 & \textbf{32.226}\\
    16 & 80 & 160 & 10 & 8 & \textbf{32.540}\\
    17 & 80 & 200 & 6 & 1 & \textbf{18.736}\\
    18 & 80 & 200 & 6 & 2 & \textbf{24.737}\\
    19 & 80 & 200 & 6 & 4 & \textbf{29.091}\\
    20 & 80 & 200 & 6 & 6 & \textbf{31.641}\\
    21 & 80 & 200 & 6 & 8 & \textbf{33.014}\\
    22 & 80 & 200 & 8 & 1 & \textbf{18.828}\\
    23 & 80 & 200 & 8 & 2 & \textbf{24.813}\\
    24 & 80 & 200 & 8 & 4 & \textbf{30.452}\\
    25 & 80 & 200 & 8 & 6 & \textbf{32.649}\\
    26 & 80 & 200 & 8 & 8 & \textbf{31.884}\\
    27 & 80 & 200 & 10 & 1 & \textbf{18.865}\\
    28 & 80 & 200 & 10 & 2 & \textbf{24.675}\\
    29 & 80 & 200 & 10 & 4 & \textbf{30.363}\\
    30 & 80 & 200 & 10 & 6 & \textbf{32.716}\\
    31 & 80 & 200 & 10 & 8 & \textbf{33.072}\\
    32 & 80 & 256 & 6 & 1 & \textbf{21.166}\\
    33 & 80 & 256 & 6 & 2 & \textbf{26.052}\\
    34 & 80 & 256 & 6 & 4 & \textbf{30.279}\\
    35 & 80 & 256 & 6 & 6 & \textbf{32.325}\\
    36 & 80 & 256 & 6 & 8 & \textbf{32.812}\\
    37 & 80 & 256 & 8 & 1 & \textbf{20.683}\\
    38 & 80 & 256 & 8 & 2 & \textbf{27.123}\\
    39 & 80 & 256 & 8 & 4 & \textbf{31.829}\\
    40 & 80 & 256 & 8 & 6 & \textbf{33.622}\\
    41 & 80 & 256 & 8 & 8 & \textbf{33.247}\\
    42 & 80 & 256 & 10 & 1 & \textbf{20.546}\\
    43 & 80 & 256 & 10 & 2 & \textbf{26.987}\\
    44 & 80 & 256 & 10 & 4 & \textbf{30.679}\\
    45 & 80 & 256 & 10 & 6 & \textbf{34.665}\\
    46 & 80 & 256 & 10 & 8 & \textbf{33.445}\\
    \bottomrule
    \end{tabular}
\end{table}

\begin{table}[htbp]
    \centering
    \setlength{\tabcolsep}{3pt} % 同样减小列间距
    \caption{Impact of policy on Mixtral 8x7B in Env \#1, C-Eval dataset.}
    \label{tab:appendix_8x7B_policy_ceval}
    \begin{tabular}{>{\centering\arraybackslash}m{1cm} 
                    >{\centering\arraybackslash}m{2.5cm} 
                    >{\centering\arraybackslash}m{2.5cm}
                    >{\centering\arraybackslash}m{2.5cm}
                    >{\centering\arraybackslash}m{2.5cm}
                    >{\centering\arraybackslash}m{2cm}}
    \toprule
         No. & Prefill Batch Size & Decoding Batch Size & Draft batch size & Draft max new token & \textbf{Throughput (token/s)}\\
    \midrule
        1 & 96 & 256 & 8 & 4 & \textbf{26.489}\\
        2 & 96 & 288 & 8 & 4 & \textbf{28.449}\\
        3 & 96 & 300 & 8 & 4 & \textbf{28.209}\\
        4 & 96 & 256 & 6 & 2 & \textbf{25.363}\\
        5 & 96 & 256 & 6 & 4 & \textbf{27.823}\\
        6 & 96 & 256 & 6 & 6 & \textbf{28.712}\\
        7 & 96 & 256 & 6 & 8 & \textbf{28.531}\\
        8 & 96 & 256 & 8 & 2 & \textbf{25.347}\\
        9 & 96 & 256 & 8 & 4 & \textbf{27.449}\\
        10 & 96 & 288 & 6 & 2 & \textbf{25.254}\\
        11 & 96 & 288 & 6 & 4 & \textbf{28.685}\\
        12 & 96 & 288 & 6 & 6 & \textbf{29.199}\\
        13 & 96 & 288 & 6 & 8 & \textbf{29.385}\\
        14 & 96 & 288 & 8 & 2 & \textbf{26.126}\\
        15 & 96 & 288 & 8 & 4 & \textbf{28.679}\\
        16 & 96 & 288 & 8 & 6 & \textbf{29.329}\\
        17 & 96 & 300 & 6 & 2 & \textbf{24.821}\\
        18 & 96 & 300 & 6 & 4 & \textbf{28.240}\\
        19 & 96 & 300 & 6 & 6 & \textbf{29.134}\\
        20 & 96 & 300 & 6 & 8 & \textbf{30.781}\\
        21 & 96 & 300 & 8 & 2 & \textbf{26.268}\\
        22 & 96 & 300 & 8 & 4 & \textbf{30.652}\\
        23 & 96 & 300 & 8 & 6 & \textbf{31.968}\\
    \bottomrule
    \end{tabular}
\end{table}

\begin{table}[htbp]
    \centering
    \setlength{\tabcolsep}{3pt} % 同样减小列间距
    \caption{Impact of policy on Mixtral 8x7B in Env \#1, SummEval dataset.}
    \label{tab:appendix_8x7B_policy_summeval}
    \begin{tabular}{>{\centering\arraybackslash}m{1cm} 
                    >{\centering\arraybackslash}m{2.5cm} 
                    >{\centering\arraybackslash}m{2.5cm}
                    >{\centering\arraybackslash}m{2.5cm}
                    >{\centering\arraybackslash}m{2.5cm}
                    >{\centering\arraybackslash}m{2cm}}
    \toprule
         No. & Prefill Batch Size & Decoding Batch Size & Draft batch size & Draft max new token & \textbf{Throughput (token/s)}\\
    \midrule
    1 & 50 & 128 & 5 & 5 & \textbf{18.937}\\
    2 & 50 & 128 & 5 & 3 & \textbf{19.735}\\
    3 & 50 & 256 & 5 & 5 & \textbf{19.890}\\
    4 & 50 & 256 & 5 & 3 & \textbf{17.560}\\
    5 & 50 & 256 & 5 & 2 & \textbf{15.624}\\
    6 & 80 & 128 & 5 & 1 & \textbf{11.682}\\
    7 & 80 & 128 & 5 & 2 & \textbf{14.509}\\
    8 & 80 & 128 & 5 & 4 & \textbf{19.464}\\
    9 & 80 & 128 & 5 & 6 & \textbf{21.166}\\
    10 & 80 & 128 & 5 & 8 & \textbf{21.531}\\
    11 & 80 & 128 & 8 & 1 & \textbf{11.629}\\
    12 & 80 & 128 & 8 & 2 & \textbf{14.408}\\
    13 & 80 & 128 & 8 & 4 & \textbf{18.321}\\
    14 & 80 & 128 & 8 & 6 & \textbf{16.989}\\
    15 & 80 & 128 & 8 & 8 & \textbf{21.958}\\
    16 & 80 & 192 & 5 & 1 & \textbf{14.764}\\
    17 & 80 & 192 & 5 & 2 & \textbf{16.830}\\
    18 & 80 & 192 & 5 & 4 & \textbf{21.072}\\
    19 & 80 & 192 & 5 & 6 & \textbf{22.029}\\
    20 & 80 & 192 & 5 & 8 & \textbf{22.712}\\
    21 & 80 & 192 & 8 & 1 & \textbf{14.305}\\
    22 & 80 & 192 & 8 & 2 & \textbf{16.757}\\
    23 & 80 & 192 & 8 & 4 & \textbf{21.435}\\
    24 & 80 & 192 & 8 & 6 & \textbf{23.653}\\
    25 & 80 & 192 & 8 & 8 & \textbf{24.732}\\
    26 & 80 & 256 & 5 & 1 & \textbf{14.809}\\
    27 & 80 & 256 & 5 & 2 & \textbf{16.781}\\
    28 & 80 & 256 & 5 & 4 & \textbf{20.441}\\
    29 & 80 & 256 & 5 & 6 & \textbf{21.841}\\
    30 & 80 & 256 & 5 & 8 & \textbf{21.741}\\
    31 & 80 & 256 & 8 & 1 & \textbf{13.822}\\
    32 & 80 & 256 & 8 & 2 & \textbf{16.265}\\
    33 & 80 & 256 & 8 & 4 & \textbf{17.243}\\
    34 & 80 & 256 & 8 & 6 & \textbf{12.903}\\
    35 & 80 & 256 & 8 & 8 & \textbf{11.103}\\
    36 & 80 & 320 & 5 & 1 & \textbf{4.444}\\
    37 & 80 & 320 & 5 & 2 & \textbf{5.757}\\
    38 & 80 & 320 & 5 & 4 & \textbf{7.761}\\
    39 & 80 & 320 & 5 & 6 & \textbf{12.376}\\
    40 & 80 & 320 & 5 & 8 & \textbf{11.503}\\
    41 & 80 & 320 & 8 & 1 & \textbf{4.550}\\
    42 & 80 & 320 & 8 & 2 & \textbf{6.074}\\
    43 & 80 & 320 & 8 & 4 & \textbf{11.785}\\
    44 & 80 & 320 & 8 & 6 & \textbf{13.218}\\
    45 & 80 & 320 & 8 & 8 & \textbf{11.293}\\
    \bottomrule
    \end{tabular}
\end{table}

\begin{table}[htbp]
    \centering
    \setlength{\tabcolsep}{3pt} % 同样减小列间距
    \caption{Impact of policy on Mixtral 8x22B in Env \#2, HumanEval dataset.}
    \label{tab:appendix_8x22B_policy_humaneval}
    \begin{tabular}{>{\centering\arraybackslash}m{1cm} 
                    >{\centering\arraybackslash}m{2.5cm} 
                    >{\centering\arraybackslash}m{2.5cm}
                    >{\centering\arraybackslash}m{2.5cm}
                    >{\centering\arraybackslash}m{2.5cm}
                    >{\centering\arraybackslash}m{2cm}}
    \toprule
         No. & Prefill Batch Size & Decoding Batch Size & Draft batch size & Draft max new token & \textbf{Throughput (token/s)}\\
    \midrule
    1 & 32 & 128 & 4 & 4 & \textbf{7.112}\\
    2 & 32 & 128 & 4 & 6 & \textbf{7.921}\\
    3 & 32 & 128 & 4 & 8 & \textbf{7.564}\\
    4 & 32 & 128 & 6 & 4 & \textbf{8.617}\\
    5 & 32 & 128 & 6 & 6 & \textbf{7.901}\\
    6 & 32 & 128 & 6 & 8 & \textbf{7.155}\\
    7 & 32 & 128 & 8 & 4 & \textbf{6.599}\\
    8 & 32 & 128 & 8 & 6 & \textbf{7.913}\\
    9 & 32 & 128 & 8 & 8 & \textbf{7.677}\\
    10 & 32 & 192 & 4 & 4 & \textbf{7.291}\\
    11 & 32 & 192 & 4 & 6 & \textbf{7.083}\\
    12 & 32 & 192 & 4 & 8 & \textbf{4.874}\\
    13 & 32 & 192 & 6 & 4 & \textbf{7.753}\\
    14 & 32 & 192 & 6 & 4 & \textbf{7.733}\\
    15 & 32 & 192 & 6 & 6 & \textbf{7.578}\\
    16 & 32 & 192 & 6 & 8 & \textbf{4.510}\\
    17 & 32 & 192 & 8 & 4 & \textbf{8.536}\\
    18 & 32 & 192 & 8 & 6 & \textbf{6.574}\\
    \bottomrule
    \end{tabular}
\end{table}

\begin{table}[htbp]
    \centering
    \setlength{\tabcolsep}{3pt} % 同样减小列间距
    \caption{Impact of policy on Mixtral 8x22B in Env \#2, C-Eval dataset.}
    \label{tab:appendix_8x22B_policy_ceval}
    \begin{tabular}{>{\centering\arraybackslash}m{1cm} 
                    >{\centering\arraybackslash}m{2.5cm} 
                    >{\centering\arraybackslash}m{2.5cm}
                    >{\centering\arraybackslash}m{2.5cm}
                    >{\centering\arraybackslash}m{2.5cm}
                    >{\centering\arraybackslash}m{2cm}}
    \toprule
         No. & Prefill Batch Size & Decoding Batch Size & Draft batch size & Draft max new token & \textbf{Throughput (token/s)}\\
    \midrule
    1 & 16 & 32 & 6 & 4 & \textbf{3.430}\\
    2 & 16 & 32 & 6 & 6 & \textbf{4.510}\\
    3 & 16 & 32 & 6 & 8 & \textbf{4.321}\\
    4 & 16 & 32 & 8 & 4 & \textbf{3.607}\\
    5 & 16 & 32 & 8 & 6 & \textbf{4.230}\\
    6 & 16 & 32 & 8 & 8 & \textbf{4.742}\\
    7 & 32 & 32 & 6 & 4 & \textbf{3.726}\\
    8 & 32 & 32 & 6 & 6 & \textbf{4.977}\\
    9 & 32 & 32 & 6 & 8 & \textbf{4.513}\\
    10 & 32 & 32 & 8 & 4 & \textbf{3.969}\\
    11 & 32 & 32 & 8 & 6 & \textbf{4.233}\\
    12 & 32 & 32 & 8 & 8 & \textbf{3.894}\\
    13 & 32 & 32 & 6 & 4 & \textbf{3.543}\\
    14 & 32 & 32 & 6 & 6 & \textbf{4.647}\\
    15 & 32 & 32 & 6 & 8 & \textbf{4.063}\\
    16 & 32 & 32 & 8 & 4 & \textbf{4.030}\\
    17 & 32 & 32 & 8 & 6 & \textbf{4.231}\\
    18 & 32 & 32 & 8 & 8 & \textbf{3.609}\\
    19 & 16 & 64 & 6 & 4 & \textbf{4.160}\\
    20 & 16 & 64 & 6 & 6 & \textbf{4.510}\\
    21 & 16 & 64 & 6 & 8 & \textbf{3.915}\\
    22 & 16 & 64 & 8 & 4 & \textbf{3.588}\\
    \bottomrule
    \end{tabular}
\end{table}

\begin{table}[htbp]
    \centering
    \setlength{\tabcolsep}{3pt} % 同样减小列间距
    \caption{Impact of policy on Mixtral 8x22B in Env \#2, SummEval dataset.}
    \label{tab:appendix_8x22B_policy_summeval}
    \begin{tabular}{>{\centering\arraybackslash}m{1cm} 
                    >{\centering\arraybackslash}m{2.5cm} 
                    >{\centering\arraybackslash}m{2.5cm}
                    >{\centering\arraybackslash}m{2.5cm}
                    >{\centering\arraybackslash}m{2.5cm}
                    >{\centering\arraybackslash}m{2cm}}
    \toprule
         No. & Prefill Batch Size & Decoding Batch Size & Draft batch size & Draft max new token & \textbf{Throughput (token/s)}\\
    \midrule
    1 & 16 & 64 & 6 & 4 & \textbf{4.579}\\
    2 & 16 & 32 & 6 & 4 & \textbf{3.711}\\
    3 & 16 & 32 & 6 & 6 & \textbf{3.486}\\
    4 & 16 & 32 & 6 & 8 & \textbf{4.225}\\
    5 & 16 & 32 & 8 & 4 & \textbf{3.862}\\
    6 & 16 & 32 & 8 & 6 & \textbf{3.998}\\
    7 & 16 & 32 & 8 & 8 & \textbf{3.975}\\
    8 & 16 & 64 & 6 & 4 & \textbf{4.529}\\
    9 & 16 & 64 & 6 & 6 & \textbf{5.141}\\
    10 & 16 & 64 & 6 & 8 & \textbf{4.977}\\
    11 & 16 & 64 & 8 & 4 & \textbf{4.546}\\
    12 & 16 & 64 & 8 & 6 & \textbf{4.590}\\
    13 & 16 & 64 & 8 & 8 & \textbf{5.911}\\
    \bottomrule
    \end{tabular}
\end{table}

\subsubsection{GPU Memory Usage}
\label{Appendix:vram}

\begin{figure}[htbp]
    \centering
    \includegraphics[width=\linewidth]{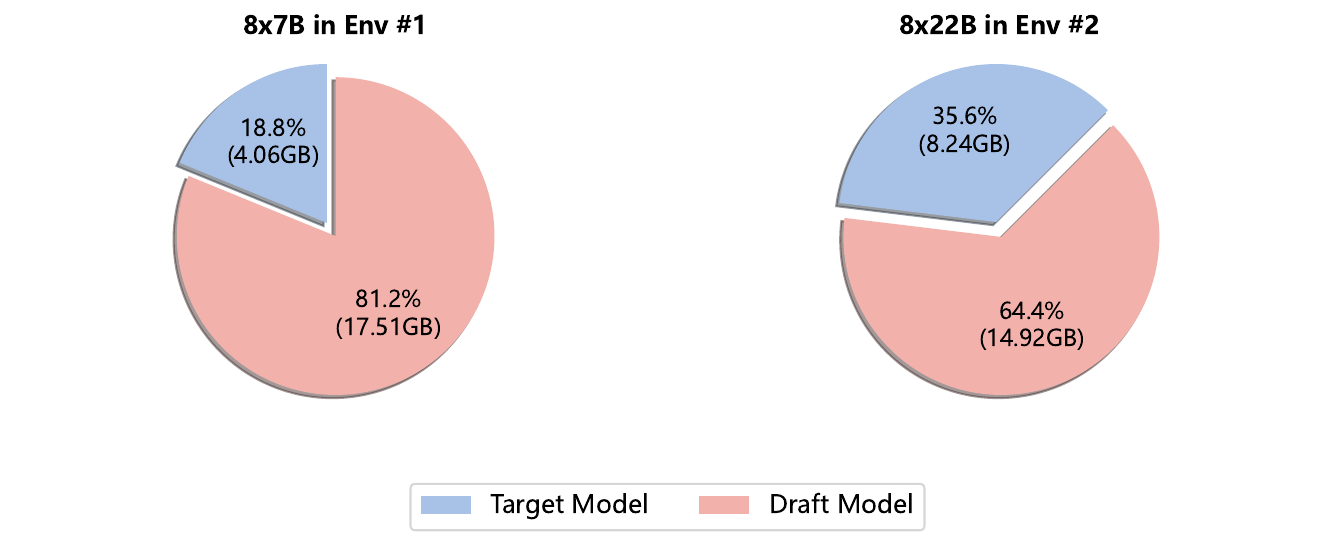}
    \caption{GPU Memory Allocation Overview.}
    \label{fig:gpu_pie}
\end{figure}
\vspace{3em}

\begin{figure}[htbp]
    \centering
    \includegraphics[width=\linewidth]{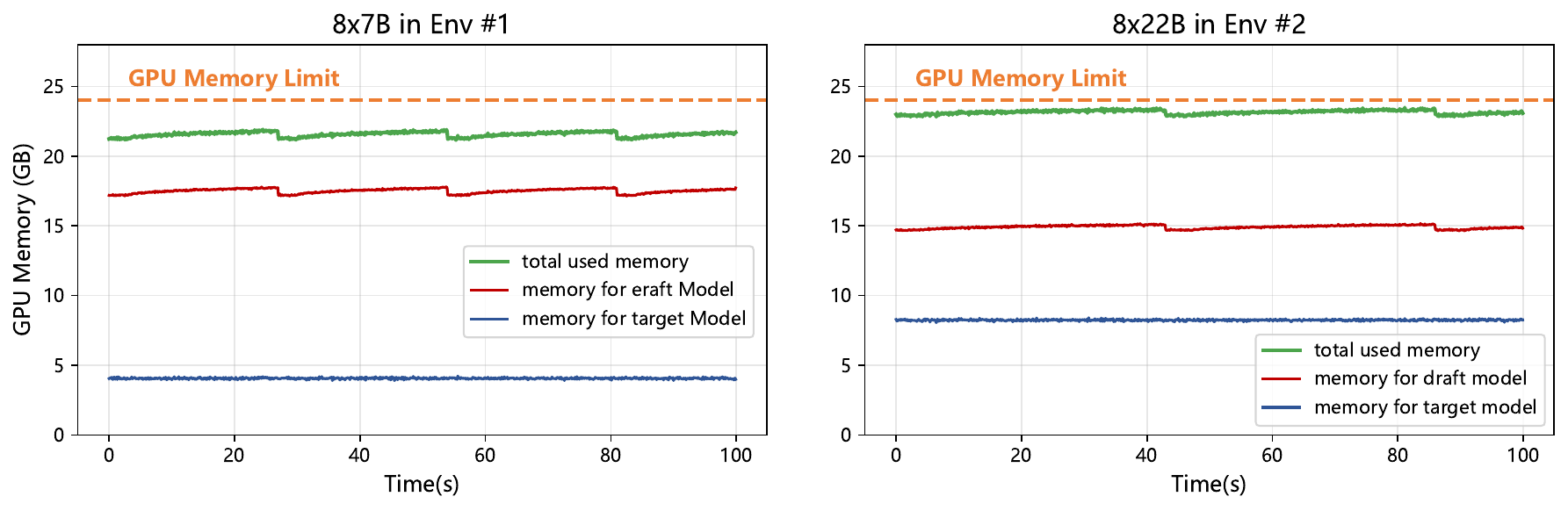}
    \caption{Runtime GPU Memory Monitoring.}
    \label{fig:gpu_line}
\end{figure}
\vspace{1em}

We used NVIDIA Nsight \cite{nsight} to monitor runtime GPU memory usage on the SummEval dataset. As shown in \figref{fig:gpu_pie}, only the parameters essential for target model offloading are retained in memory, while the remaining space is occupied by the draft model and its cache. This aligns with our design rationale: during offloading, it is more efficient to allocate GPU memory to the draft model rather than storing the target model parameters.

Runtime GPU memory monitoring reveals a periodic pattern in the draft model’s memory usage. As shown in the left panel of \figref{fig:gpu_line}, each cycle lasts approximately 28 seconds, characterized by a gradual increase in memory usage followed by a sharp drop and a 2-second idle window. This aligns with the behavior observed in \figref{fig:gpu-util}, where the draft model performs computation for 26 seconds and remains idle for 2 seconds awaiting the next batch.

\subsubsection{Ablation Study}
\label{Appendix:ablation}

In addition to the main results, we conducted ablation studies on other datasets. The results are as follows:

\vspace{2em}
\begin{table}[htbp]
    \centering
    \setlength{\tabcolsep}{3pt} % 同样减小列间距
    \caption{Ablation study of proposed techniques on HumanEval dataset. The gray tuple denotes a policy (Prefill batch size, decoding batch size, draft batch size, draft max new tokens).}
    \label{tab:ablation_humaneval}
    \begin{tabular}{>{\centering\arraybackslash}m{0.8cm} 
                    >{\centering\arraybackslash}m{3.2cm} 
                    >{\centering\arraybackslash}m{3.0cm}
                    >{\centering\arraybackslash}m{3.2cm} 
                    >{\centering\arraybackslash}m{3cm}}
    \toprule
     & All optimizations & No policy search & Serial SD & No SD \\
    \midrule
    8x7B & 34.665 \textcolor{gray}{(80, 256, 10, 6)}& 15.869  \textcolor{gray}{(80, 160, 6, 1)} & 15.005 \textcolor{gray}{(80, 256, 32, 6)} & 16.468 \textcolor{gray}{(80, 192, x, x)}\\
    8x22B & 8.617 \textcolor{gray}{(32, 128, 6, 4)}  & 4.510 \textcolor{gray}{(32, 192, 6, 8)}& 5.264 \textcolor{gray}{(32, 128, 32, 8)} & 4.108 \textcolor{gray}{(16, 64, x, x)}\\
    \bottomrule
    \end{tabular}
\end{table}

\begin{table}[htbp]
    \centering
    \setlength{\tabcolsep}{3pt} % 同样减小列间距
    \caption{Ablation study of proposed techniques on C-Eval dataset. The gray tuple denotes a policy (Prefill batch size, decoding batch size, draft batch size, draft max new tokens).}
    \label{tab:ablation_ceval}
    \begin{tabular}{>{\centering\arraybackslash}m{0.8cm} 
                    >{\centering\arraybackslash}m{3.2cm} 
                    >{\centering\arraybackslash}m{3.0cm}
                    >{\centering\arraybackslash}m{3.2cm} 
                    >{\centering\arraybackslash}m{3cm}}
    \toprule
     & All optimizations & No policy search & Serial SD & No SD \\
    \midrule
    8x7B & 31.968 \textcolor{gray}{(96, 300, 8, 6)}& 26.126 \textcolor{gray}{(96, 288, 8, 2)} & 21.989 \textcolor{gray}{(96, 288, 24, 6)}& 15.106 \textcolor{gray}{(96, 288, x, x)} \\
    8x22B & 4.977 \textcolor{gray}{(32, 32, 6, 6)}  & 3.588 \textcolor{gray}{(16, 64, 8, 4)}& 3.820 \textcolor{gray}{(32, 64, 16, 6)}  & 1.812 \textcolor{gray}{(32, 64, x, x)} \\
    \bottomrule
    \end{tabular}
\end{table}

\begin{table}[H]
    \centering
    \setlength{\tabcolsep}{3pt} % 同样减小列间距
    \caption{Ablation study of proposed techniques on SAMSum dataset. The gray tuple denotes a policy (Prefill batch size, decoding batch size, draft batch size, draft max new tokens).}
    \label{tab:ablation_samsum}
    \begin{tabular}{>{\centering\arraybackslash}m{0.8cm} 
                    >{\centering\arraybackslash}m{3.2cm} 
                    >{\centering\arraybackslash}m{3.0cm}
                    >{\centering\arraybackslash}m{3.2cm} 
                    >{\centering\arraybackslash}m{3cm}}
    \toprule
     & All optimizations & No policy search & Serial SD & No SD \\
    \midrule
    8x7B & 21.109 \textcolor{gray}{(100, 300, 6, 4)}& 12.694 \textcolor{gray}{(80, 256, 8, 2)} & 13.64 \textcolor{gray}{(100, 300, 24, 4)} & 13.072 \textcolor{gray}{(80, 256, x, x)}\\
    8x22B & 4.139 \textcolor{gray}{(16, 64, 8, 6)}  & 3.059  \textcolor{gray}{(16, 64, 6, 4)}& 3.544 \textcolor{gray}{(16, 64, 16, 6)} & 2.378\textcolor{gray}{(16, 80, x, x)} \\
    \bottomrule
    \end{tabular}
\end{table}

\subsection{Limitation}
\label{appendix:limitation}
The main limitation of this paper lies in the fact that speculative decoding is not a consistently reliable method for acceleration. In extreme cases, none of the draft tokens in multiple batches may be accepted, which greatly limits the acceleration effect of \methodname.

%%%%%%%%%%%%%%%%%%%%%%%%%%%%%%%%%%%%%%%%%%%%%%%%%%%%%%%%%%%%
\clearpage
\newpage
\section*{NeurIPS Paper Checklist}

\begin{enumerate}

\item {\bf Claims}
    \item[] Question: Do the main claims made in the abstract and introduction accurately reflect the paper's contributions and scope?
    \item[] Answer: \answerYes{} % Replace by \answerYes{}, \answerNo{}, or \answerNA{}.
    \item[] Justification: All claims are supported by the experimental results.
    \item[] Guidelines:
    \begin{itemize}
        \item The answer NA means that the abstract and introduction do not include the claims made in the paper.
        \item The abstract and/or introduction should clearly state the claims made, including the contributions made in the paper and important assumptions and limitations. A No or NA answer to this question will not be perceived well by the reviewers. 
        \item The claims made should match theoretical and experimental results, and reflect how much the results can be expected to generalize to other settings. 
        \item It is fine to include aspirational goals as motivation as long as it is clear that these goals are not attained by the paper. 
    \end{itemize}

\item {\bf Limitations}
    \item[] Question: Does the paper discuss the limitations of the work performed by the authors?
    \item[] Answer: \answerYes{} % Replace by \answerYes{}, \answerNo{}, or \answerNA{}.
    \item[] Justification: We discuss limitations in \appendixref{appendix:limitation}.
    \item[] Guidelines:
    \begin{itemize}
        \item The answer NA means that the paper has no limitation while the answer No means that the paper has limitations, but those are not discussed in the paper. 
        \item The authors are encouraged to create a separate "Limitations" section in their paper.
        \item The paper should point out any strong assumptions and how robust the results are to violations of these assumptions (e.g., independence assumptions, noiseless settings, model well-specification, asymptotic approximations only holding locally). The authors should reflect on how these assumptions might be violated in practice and what the implications would be.
        \item The authors should reflect on the scope of the claims made, e.g., if the approach was only tested on a few datasets or with a few runs. In general, empirical results often depend on implicit assumptions, which should be articulated.
        \item The authors should reflect on the factors that influence the performance of the approach. For example, a facial recognition algorithm may perform poorly when image resolution is low or images are taken in low lighting. Or a speech-to-text system might not be used reliably to provide closed captions for online lectures because it fails to handle technical jargon.
        \item The authors should discuss the computational efficiency of the proposed algorithms and how they scale with dataset size.
        \item If applicable, the authors should discuss possible limitations of their approach to address problems of privacy and fairness.
        \item While the authors might fear that complete honesty about limitations might be used by reviewers as grounds for rejection, a worse outcome might be that reviewers discover limitations that aren't acknowledged in the paper. The authors should use their best judgment and recognize that individual actions in favor of transparency play an important role in developing norms that preserve the integrity of the community. Reviewers will be specifically instructed to not penalize honesty concerning limitations.
    \end{itemize}

\item {\bf Theory assumptions and proofs}
    \item[] Question: For each theoretical result, does the paper provide the full set of assumptions and a complete (and correct) proof?
    \item[] Answer: \answerYes{} % Replace by \answerYes{}, \answerNo{}, or \answerNA{}.
    \item[] Justification: All assumptions and complete proofs related to the Planner component are provided in the \appendixref{Appendix:planner}.
    \item[] Guidelines:
    \begin{itemize}
        \item The answer NA means that the paper does not include theoretical results. 
        \item All the theorems, formulas, and proofs in the paper should be numbered and cross-referenced.
        \item All assumptions should be clearly stated or referenced in the statement of any theorems.
        \item The proofs can either appear in the main paper or the supplemental material, but if they appear in the supplemental material, the authors are encouraged to provide a short proof sketch to provide intuition. 
        \item Inversely, any informal proof provided in the core of the paper should be complemented by formal proofs provided in appendix or supplemental material.
        \item Theorems and Lemmas that the proof relies upon should be properly referenced. 
    \end{itemize}

    \item {\bf Experimental result reproducibility}
    \item[] Question: Does the paper fully disclose all the information needed to reproduce the main experimental results of the paper to the extent that it affects the main claims and/or conclusions of the paper (regardless of whether the code and data are provided or not)?
    \item[] Answer: \answerYes{} % Replace by \answerYes{}, \answerNo{}, or \answerNA{}.
    \item[] Justification: Our results are reproducible using the attached code. The necessary setup details are given in \appendixref{Appendix: implementation}.
    \item[] Guidelines:
    \begin{itemize}
        \item The answer NA means that the paper does not include experiments.
        \item If the paper includes experiments, a No answer to this question will not be perceived well by the reviewers: Making the paper reproducible is important, regardless of whether the code and data are provided or not.
        \item If the contribution is a dataset and/or model, the authors should describe the steps taken to make their results reproducible or verifiable. 
        \item Depending on the contribution, reproducibility can be accomplished in various ways. For example, if the contribution is a novel architecture, describing the architecture fully might suffice, or if the contribution is a specific model and empirical evaluation, it may be necessary to either make it possible for others to replicate the model with the same dataset, or provide access to the model. In general. releasing code and data is often one good way to accomplish this, but reproducibility can also be provided via detailed instructions for how to replicate the results, access to a hosted model (e.g., in the case of a large language model), releasing of a model checkpoint, or other means that are appropriate to the research performed.
        \item While NeurIPS does not require releasing code, the conference does require all submissions to provide some reasonable avenue for reproducibility, which may depend on the nature of the contribution. For example
        \begin{enumerate}
            \item If the contribution is primarily a new algorithm, the paper should make it clear how to reproduce that algorithm.
            \item If the contribution is primarily a new model architecture, the paper should describe the architecture clearly and fully.
            \item If the contribution is a new model (e.g., a large language model), then there should either be a way to access this model for reproducing the results or a way to reproduce the model (e.g., with an open-source dataset or instructions for how to construct the dataset).
            \item We recognize that reproducibility may be tricky in some cases, in which case authors are welcome to describe the particular way they provide for reproducibility. In the case of closed-source models, it may be that access to the model is limited in some way (e.g., to registered users), but it should be possible for other researchers to have some path to reproducing or verifying the results.
        \end{enumerate}
    \end{itemize}

\item {\bf Open access to data and code}
    \item[] Question: Does the paper provide open access to the data and code, with sufficient instructions to faithfully reproduce the main experimental results, as described in supplemental material?
    \item[] Answer: \answerYes{} % Replace by \answerYes{}, \answerNo{}, or \answerNA{}.
    \item[] Justification: Our code is available at \href{www.baidu.com}{www.baidu.com}. All datasets and models used are publicly available.
    \item[] Guidelines:
    \begin{itemize}
        \item The answer NA means that paper does not include experiments requiring code.
        \item Please see the NeurIPS code and data submission guidelines (\url{https://nips.cc/public/guides/CodeSubmissionPolicy}) for more details.
        \item While we encourage the release of code and data, we understand that this might not be possible, so “No” is an acceptable answer. Papers cannot be rejected simply for not including code, unless this is central to the contribution (e.g., for a new open-source benchmark).
        \item The instructions should contain the exact command and environment needed to run to reproduce the results. See the NeurIPS code and data submission guidelines (\url{https://nips.cc/public/guides/CodeSubmissionPolicy}) for more details.
        \item The authors should provide instructions on data access and preparation, including how to access the raw data, preprocessed data, intermediate data, and generated data, etc.
        \item The authors should provide scripts to reproduce all experimental results for the new proposed method and baselines. If only a subset of experiments are reproducible, they should state which ones are omitted from the script and why.
        \item At submission time, to preserve anonymity, the authors should release anonymized versions (if applicable).
        \item Providing as much information as possible in supplemental material (appended to the paper) is recommended, but including URLs to data and code is permitted.
    \end{itemize}

\item {\bf Experimental setting/details}
    \item[] Question: Does the paper specify all the training and test details (e.g., data splits, hyperparameters, how they were chosen, type of optimizer, etc.) necessary to understand the results?
    \item[] Answer: \answerYes{} % Replace by \answerYes{}, \answerNo{}, or \answerNA{}.
    \item[] Justification: The paper, together with the \appendixref{Appendix: implementation} and instructions in the code, contain information on all substantial details necessary to understand the results.
    \item[] Guidelines:
    \begin{itemize}
        \item The answer NA means that the paper does not include experiments.
        \item The experimental setting should be presented in the core of the paper to a level of detail that is necessary to appreciate the results and make sense of them.
        \item The full details can be provided either with the code, in appendix, or as supplemental material.
    \end{itemize}

\item {\bf Experiment statistical significance}
    \item[] Question: Does the paper report error bars suitably and correctly defined or other appropriate information about the statistical significance of the experiments?
    \item[] Answer: \answerNo{} % Replace by \answerYes{}, \answerNo{}, or \answerNA{}.
    \item[] Justification: The runs for the experiments in the paper have low variance.
    \item[] Guidelines:
    \begin{itemize}
        \item The answer NA means that the paper does not include experiments.
        \item The authors should answer "Yes" if the results are accompanied by error bars, confidence intervals, or statistical significance tests, at least for the experiments that support the main claims of the paper.
        \item The factors of variability that the error bars are capturing should be clearly stated (for example, train/test split, initialization, random drawing of some parameter, or overall run with given experimental conditions).
        \item The method for calculating the error bars should be explained (closed form formula, call to a library function, bootstrap, etc.)
        \item The assumptions made should be given (e.g., Normally distributed errors).
        \item It should be clear whether the error bar is the standard deviation or the standard error of the mean.
        \item It is OK to report 1-sigma error bars, but one should state it. The authors should preferably report a 2-sigma error bar than state that they have a 96\% CI, if the hypothesis of Normality of errors is not verified.
        \item For asymmetric distributions, the authors should be careful not to show in tables or figures symmetric error bars that would yield results that are out of range (e.g. negative error rates).
        \item If error bars are reported in tables or plots, The authors should explain in the text how they were calculated and reference the corresponding figures or tables in the text.
    \end{itemize}

\item {\bf Experiments compute resources}
    \item[] Question: For each experiment, does the paper provide sufficient information on the computer resources (type of compute workers, memory, time of execution) needed to reproduce the experiments?
    \item[] Answer: \answerYes{} % Replace by \answerYes{}, \answerNo{}, or \answerNA{}.
    \item[] Justification: Yes, we provide sufficient information on the computer resources needed to reproduce the experiments in \secref{sec:expsetup}.
    \item[] Guidelines:
    \begin{itemize}
        \item The answer NA means that the paper does not include experiments.
        \item The paper should indicate the type of compute workers CPU or GPU, internal cluster, or cloud provider, including relevant memory and storage.
        \item The paper should provide the amount of compute required for each of the individual experimental runs as well as estimate the total compute. 
        \item The paper should disclose whether the full research project required more compute than the experiments reported in the paper (e.g., preliminary or failed experiments that didn't make it into the paper). 
    \end{itemize}
    
\item {\bf Code of ethics}
    \item[] Question: Does the research conducted in the paper conform, in every respect, with the NeurIPS Code of Ethics \url{https://neurips.cc/public/EthicsGuidelines}?
    \item[] Answer: \answerYes{} % Replace by \answerYes{}, \answerNo{}, or \answerNA{}.
    \item[] Justification:  Yes, we confirm adherence to the NeurIPS Code of Ethics.
    \item[] Guidelines:
    \begin{itemize}
        \item The answer NA means that the authors have not reviewed the NeurIPS Code of Ethics.
        \item If the authors answer No, they should explain the special circumstances that require a deviation from the Code of Ethics.
        \item The authors should make sure to preserve anonymity (e.g., if there is a special consideration due to laws or regulations in their jurisdiction).
    \end{itemize}

\item {\bf Broader impacts}
    \item[] Question: Does the paper discuss both potential positive societal impacts and negative societal impacts of the work performed?
    \item[] Answer: \answerNA{} % Replace by \answerYes{}, \answerNo{}, or \answerNA{}.
    \item[] Justification: Our work is focused on efficient LLM Offloading. We do not foresee any particular societal impacts from this work.
    \item[] Guidelines: 
    \begin{itemize}
        \item The answer NA means that there is no societal impact of the work performed.
        \item If the authors answer NA or No, they should explain why their work has no societal impact or why the paper does not address societal impact.
        \item Examples of negative societal impacts include potential malicious or unintended uses (e.g., disinformation, generating fake profiles, surveillance), fairness considerations (e.g., deployment of technologies that could make decisions that unfairly impact specific groups), privacy considerations, and security considerations.
        \item The conference expects that many papers will be foundational research and not tied to particular applications, let alone deployments. However, if there is a direct path to any negative applications, the authors should point it out. For example, it is legitimate to point out that an improvement in the quality of generative models could be used to generate deepfakes for disinformation. On the other hand, it is not needed to point out that a generic algorithm for optimizing neural networks could enable people to train models that generate Deepfakes faster.
        \item The authors should consider possible harms that could arise when the technology is being used as intended and functioning correctly, harms that could arise when the technology is being used as intended but gives incorrect results, and harms following from (intentional or unintentional) misuse of the technology.
        \item If there are negative societal impacts, the authors could also discuss possible mitigation strategies (e.g., gated release of models, providing defenses in addition to attacks, mechanisms for monitoring misuse, mechanisms to monitor how a system learns from feedback over time, improving the efficiency and accessibility of ML).
    \end{itemize}
    
\item {\bf Safeguards}
    \item[] Question: Does the paper describe safeguards that have been put in place for responsible release of data or models that have a high risk for misuse (e.g., pretrained language models, image generators, or scraped datasets)?
    \item[] Answer: \answerNA{} % Replace by \answerYes{}, \answerNo{}, or \answerNA{}.
    \item[] Justification: This paper does not alter the capabilities of the available models or datasets, but rather provides a more efficient approach to use them on hardware with limited capabilities. Thus, we believe that our paper has a neutral risk impact in this area
    \item[] Guidelines:
    \begin{itemize}
        \item The answer NA means that the paper poses no such risks.
        \item Released models that have a high risk for misuse or dual-use should be released with necessary safeguards to allow for controlled use of the model, for example by requiring that users adhere to usage guidelines or restrictions to access the model or implementing safety filters. 
        \item Datasets that have been scraped from the Internet could pose safety risks. The authors should describe how they avoided releasing unsafe images.
        \item We recognize that providing effective safeguards is challenging, and many papers do not require this, but we encourage authors to take this into account and make a best faith effort.
    \end{itemize}

\item {\bf Licenses for existing assets}
    \item[] Question: Are the creators or original owners of assets (e.g., code, data, models), used in the paper, properly credited and are the license and terms of use explicitly mentioned and properly respected?
    \item[] Answer: \answerYes{} % Replace by \answerYes{}, \answerNo{}, or \answerNA{}.
    \item[] Justification: All datasets and models used  are open-source and publicly available.
    \item[] Guidelines:
    \begin{itemize}
        \item The answer NA means that the paper does not use existing assets.
        \item The authors should cite the original paper that produced the code package or dataset.
        \item The authors should state which version of the asset is used and, if possible, include a URL.
        \item The name of the license (e.g., CC-BY 4.0) should be included for each asset.
        \item For scraped data from a particular source (e.g., website), the copyright and terms of service of that source should be provided.
        \item If assets are released, the license, copyright information, and terms of use in the package should be provided. For popular datasets, \url{paperswithcode.com/datasets} has curated licenses for some datasets. Their licensing guide can help determine the license of a dataset.
        \item For existing datasets that are re-packaged, both the original license and the license of the derived asset (if it has changed) should be provided.
        \item If this information is not available online, the authors are encouraged to reach out to the asset's creators.
    \end{itemize}

\item {\bf New assets}
    \item[] Question: Are new assets introduced in the paper well documented and is the documentation provided alongside the assets?
    \item[] Answer: \answerYes{} % Replace by \answerYes{}, \answerNo{}, or \answerNA{}.
    \item[] Justification: We provide a link to the anonymized code, which has documentation provided that explains how to run the experiments.
    \item[] Guidelines:
    \begin{itemize}
        \item The answer NA means that the paper does not release new assets.
        \item Researchers should communicate the details of the dataset/code/model as part of their submissions via structured templates. This includes details about training, license, limitations, etc. 
        \item The paper should discuss whether and how consent was obtained from people whose asset is used.
        \item At submission time, remember to anonymize your assets (if applicable). You can either create an anonymized URL or include an anonymized zip file.
    \end{itemize}

\item {\bf Crowdsourcing and research with human subjects}
    \item[] Question: For crowdsourcing experiments and research with human subjects, does the paper include the full text of instructions given to participants and screenshots, if applicable, as well as details about compensation (if any)? 
    \item[] Answer: \answerNA{} % Replace by \answerYes{}, \answerNo{}, or \answerNA{}.
    \item[] Justification: We did not conduct any crowdsourcing experiments or research with human subjects.
    \item[] Guidelines:
    \begin{itemize}
        \item The answer NA means that the paper does not involve crowdsourcing nor research with human subjects.
        \item Including this information in the supplemental material is fine, but if the main contribution of the paper involves human subjects, then as much detail as possible should be included in the main paper. 
        \item According to the NeurIPS Code of Ethics, workers involved in data collection, curation, or other labor should be paid at least the minimum wage in the country of the data collector. 
    \end{itemize}

\item {\bf Institutional review board (IRB) approvals or equivalent for research with human subjects}
    \item[] Question: Does the paper describe potential risks incurred by study participants, whether such risks were disclosed to the subjects, and whether Institutional Review Board (IRB) approvals (or an equivalent approval/review based on the requirements of your country or institution) were obtained?
    \item[] Answer: \answerNA{} % Replace by \answerYes{}, \answerNo{}, or \answerNA{}.
    \item[] Justification: We did not conduct any research with human subjects.
    \item[] Guidelines:
    \begin{itemize}
        \item The answer NA means that the paper does not involve crowdsourcing nor research with human subjects.
        \item Depending on the country in which research is conducted, IRB approval (or equivalent) may be required for any human subjects research. If you obtained IRB approval, you should clearly state this in the paper. 
        \item We recognize that the procedures for this may vary significantly between institutions and locations, and we expect authors to adhere to the NeurIPS Code of Ethics and the guidelines for their institution. 
        \item For initial submissions, do not include any information that would break anonymity (if applicable), such as the institution conducting the review.
    \end{itemize}

\item {\bf Declaration of LLM usage}
    \item[] Question: Does the paper describe the usage of LLMs if it is an important, original, or non-standard component of the core methods in this research? Note that if the LLM is used only for writing, editing, or formatting purposes and does not impact the core methodology, scientific rigorousness, or originality of the research, declaration is not required.
    %this research? 
    \item[] Answer: \answerNA{} % Replace by \answerYes{}, \answerNo{}, or \answerNA{}.
    \item[] Justification: LLM is used only for writing, editing, or formatting purposes.
    \item[] Guidelines:
    \begin{itemize}
        \item The answer NA means that the core method development in this research does not involve LLMs as any important, original, or non-standard components.
        \item Please refer to our LLM policy (\url{https://neurips.cc/Conferences/2025/LLM}) for what should or should not be described.
    \end{itemize}

\end{enumerate}

\end{document}